\documentclass{article}

    \usepackage[preprint, nonatbib]{neurips_2021}

\usepackage[utf8]{inputenc} 
\usepackage[T1]{fontenc}    
\usepackage{url}            
\usepackage{booktabs}       
\usepackage{amsfonts}       
\usepackage{nicefrac}       
\usepackage{microtype}      
\usepackage{xcolor}         
\usepackage{multicol}

\usepackage{times}
\usepackage{epsfig}
\usepackage{graphicx}
\usepackage{wrapfig}
\usepackage{amsmath}
\usepackage{amssymb}
\usepackage{enumitem}

\usepackage{float}
\usepackage{cancel}
\usepackage{listings}
\usepackage{tabularx}
\usepackage{multirow}
\usepackage[numbers]{natbib}
\usepackage[style=base]{caption}
\usepackage{amsmath,amsfonts,bm}

\def\eqref#1{equation~\ref{#1}}

\def\1{\bm{1}}

\def\rva{{\mathbf{a}}}
\def\rvb{{\mathbf{b}}}
\def\rvc{{\mathbf{c}}}

\def\rvu{{\mathbf{i}}}

\def\rvu{{\mathbf{u}}}
\def\rvv{{\mathbf{v}}}

\def\rvx{{\mathbf{x}}}
\def\rvy{{\mathbf{y}}}
\def\rvz{{\mathbf{z}}}

\def\rmM{{\mathbf{M}}}

\DeclareMathAlphabet{\mathsfit}{\encodingdefault}{\sfdefault}{m}{sl}
\SetMathAlphabet{\mathsfit}{bold}{\encodingdefault}{\sfdefault}{bx}{n}

\usepackage{url}

\usepackage{color}
\usepackage{graphicx}
\usepackage{float}
\usepackage{subcaption}
\usepackage{multirow}
\usepackage{siunitx}
\newcommand{\Abs}[1]{\left\vert#1\right\vert}

\newcommand{\bq}{\begin{equation}}
\newcommand{\eq}{\end{equation}}

\newcommand{\D}{{\mathrm{d}}}

\newcommand{\barx}{\bar x}
\newcommand{\xbar}{\barx}

\newcommand\numberthis{\addtocounter{equation}{1}\tag{\theequation}}

\newcommand{\x}{{\mkern-2mu\times\mkern-2mu}}

\allowdisplaybreaks

\newcommand\Tstrut{\rule{0pt}{2.1ex}}         

\usepackage[acronym]{glossaries}

\usepackage[breaklinks=true,bookmarks=false]{hyperref}
\usepackage{cleveref}

\title{Multi-Resolution Continuous Normalizing Flows}

\author{
Vikram Voleti \\
Mila, Universit\'e de Montr\'eal \\
\and
\textbf{Chris Finlay} \\
McGill University \& Deep Render \\
\and
\textbf{Adam Oberman} \\
McGill University, Mila \\
\and
\textbf{Christopher Pal} \\
Polytechnique Montr\'eal, Mila \\
Canada CIFAR AI Chair \\
}

\begin{document}

\maketitle

\newacronym{cnf}{CNF}{Continuous Normalizing Flows}
\newacronym{gan}{GAN}{Generative Adversarial Networks}
\newacronym{msflow}{MRCNF}{Multi-Resolution Continuous Normalizing Flow}
\newacronym{msflow-image}{MRCNF}{Multi-Resolution Continuous Normalizing Flow}
\newacronym{msflow-im}{MRCNF}{Multi-Resolution Continuous Normalizing Flow}
\newacronym{msflow-wv}{MRCNF-Wavelet}{Multi-Resolution Continuous Normalizing Flow - Wavelet}
\newacronym{bpd}{BPD}{Bits-per-dimension}
\newacronym{mrf}{MRF}{Markov Random Field}
\newacronym{fid}{FID}{Fr\'echet Inception Distance}
\newacronym{ood}{OoD}{out-of-distribution}
\newacronym{mrcnf}{MRCNF}{Multi-Resolution Continuous Normalizing Flows}

\begin{abstract}
Recent work has shown that Neural Ordinary Differential Equations (ODEs) can serve as generative models of images using the perspective of Continuous Normalizing Flows (CNFs). Such models offer exact likelihood calculation, and invertible generation/density estimation. In this work we introduce a Multi-Resolution variant of such models (MRCNF), by characterizing the conditional distribution over the additional information required to generate a fine image that is consistent with the coarse image. We introduce a transformation between resolutions that allows for no change in the log likelihood. We show that this approach yields comparable likelihood values for various image datasets, with improved performance at higher resolutions, with fewer parameters, using only 1 GPU. Further, we examine the out-of-distribution properties of MRCNFs, and find that they are similar to those of other likelihood-based generative models.

\end{abstract}

\section{Introduction}
\label{section:introduction}
Reversible generative models derived through the use of the change of variables technique~\citep{dinh2016density, kingma2018glow, ho2019flow++, yu2020waveletflow} are growing in interest as alternatives to generative models based on Generative Adversarial Networks (GANs)~\citep{goodfellow2016deep} and Variational Autoencoders (VAEs)~\citep{kingma2013auto}. While GANs and VAEs have been able to produce visually impressive samples of images, they have a number of limitations. A change of variables approach facilitates the transformation of a simple base probability distribution into a more complex model distribution. Reversible generative models using this technique are attractive because they enable efficient density estimation, efficient sampling, and computation of exact likelihoods.

A promising variation of the change-of-variable approach is based on the use of a continuous time variant of normalizing flows \citep{chen2018neural,grathwohl2019ffjord, finlay2020rnode}, which uses an integral over continuous time dynamics to transform a base distribution into the model distribution, called \gls{cnf}. This approach uses ordinary differential equations (ODEs) specified by a neural network, or Neural ODEs. \gls{cnf}s have been shown to be capable of modelling complex distributions such as those associated with images.

\begin{figure}[!thb]
\vspace{-.25cm}
\begin{center}
\includegraphics[width=0.75\linewidth]{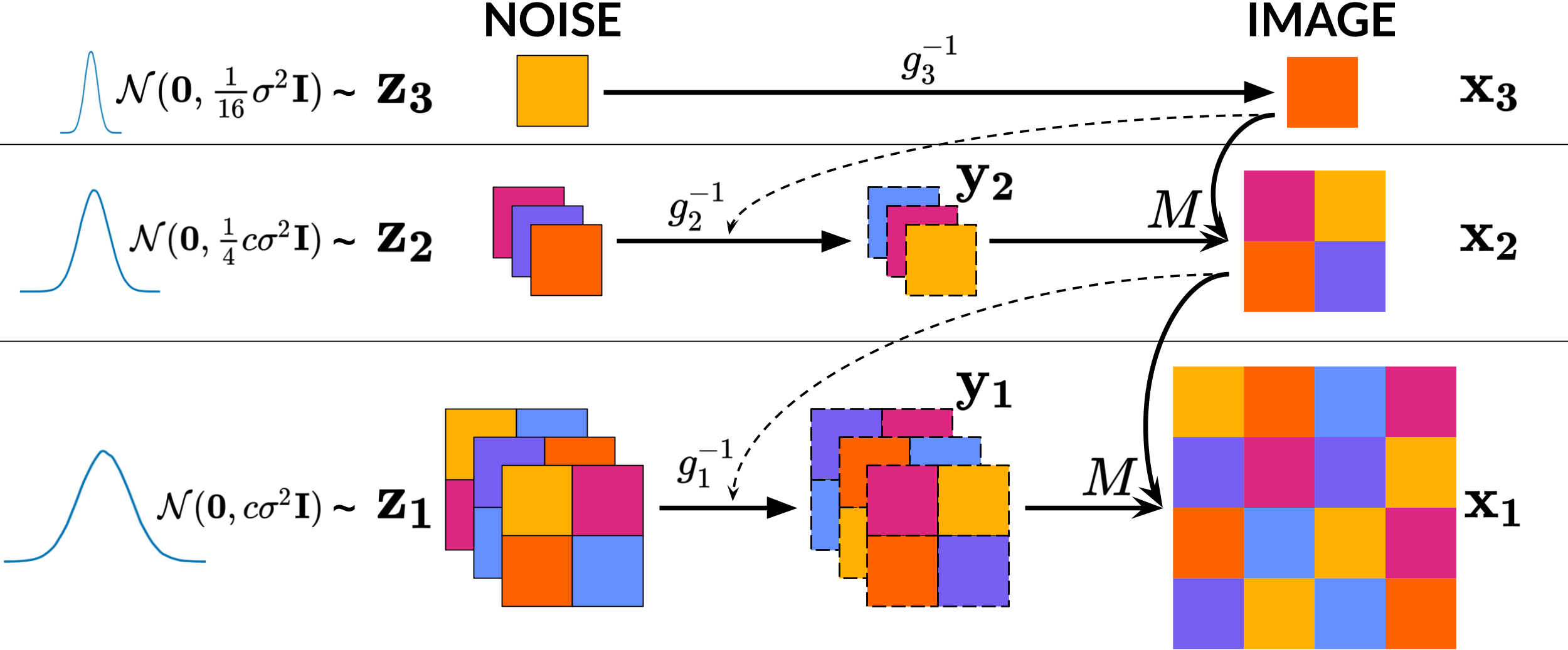}
\end{center}
\caption{The architecture of our \gls{msflow-image} method (best viewed in color). Continuous normalizing flows (CNFs) $g_s$ are used to generate images $\rvx_s$ from noise $\rvz_s$ at each resolution, with those at finer resolutions conditioned (dashed lines) on the coarser image one level above $\rvx_{s+1}$, except at the coarsest level where it is unconditional. Every finer CNF produces an intermediate image $\rvy_s$, which is then combined with the immediate coarser image $\rvx_{s+1}$ using a linear map $M$ from \cref{eq:M} to form $\rvx_s$. The multiscale maps are defined by \cref{eq:gen_cnf}.
}
\label{fig:mrcnf_2d_gen}
\end{figure}

While this new paradigm for the generative modelling of images is not as mature as GANs or VAEs in terms of the generated image quality, it is a promising direction of research as it does not have some key shortcomings associated with GANs and VAEs. Specifically, GANs are known to suffer from mode-collapse~\citep{lin2018pacgan}, and are notoriously difficult to train~\citep{arjovsky2017towards} compared to feed forward networks because their adversarial loss seeks a saddle point instead of a local minimum~\citep{berard2019closer}. \gls{cnf}s are trained by mapping images to noise, and their reversible architecture allows images to be generated by going in reverse, from noise to images. This leads to fewer issues related to mode collapse, since any input example in the dataset can be recovered from the flow 
using the reverse of the transformation learned during training. VAEs only provide a lower bound on the marginal likelihood whereas \gls{cnf}s provide exact likelihoods. Despite the many advantages of reversible generative models built with \gls{cnf}s, quantitatively such methods still do not match the widely used Fréchet Inception Distance (FID) scores of GANs or VAEs. However their other advantages motivate us to explore them further.

Furthermore, state-of-the art GANs and VAEs exploit the multi-resolution properties of images, and recent top-performing methods also inject noise at each resolution~\citep{brock2019biggan,shaham2019singan,karras2020analyzing,vahdat2020nvae}. While shaping noise is fundamental to normalizing flows, only recently have normalizing flows exploited the multi-resolution properties of images. For example, WaveletFlow~\citep{yu2020waveletflow} splits an image into multiple resolutions using the Discrete Wavelet Transform, and models the average image at each resolution using a normalizing flow. While this method has advantages, it suffers from many issues such as high parameter count and long training time.

In this work, we consider a non-trivial multi-resolution approach to continuous normalizing flows, which fixes many of these issues.
A high-level view of our approach is shown in \autoref{fig:mrcnf_2d_gen}.
%
%
Our main contributions are:
\begin{enumerate}
    \item We propose a multi-resolution transformation that does not add cost in terms of likelihood.
    \item We introduce \textbf{\gls{mrcnf}}.
    \item We achieve comparable \gls{bpd} (negative log likelihood per pixel) on image datasets using fewer model parameters and significantly less training time with only one GPU.
    \item We explore the out-of-distribution properties of (MR)CNF, and find that they are similar to non-continuous normalizing flows.
\end{enumerate}

\section{Background}
\label{background}
\subsection{Normalizing Flows}

Normalizing flows~\citep{tabak2013family, jimenez2015variational, dinh2016density, papamakarios2019normalizing, kobyzev2020normalizing} are generative models that map a complex data distribution, such as real images, to a known noise distribution. They are trained by maximizing the log likelihood of their input images. Suppose a normalizing flow $g$ produces output $\rvz$ from an input $\rvx$ i.e. $\rvz = g(\rvx)$. The change-of-variables formula provides the likelihood of the image under this transformation as:
\begin{align}
\label{eq:change_of_variables}
    \log p(\rvx) = \log \Abs{\det\frac{\D g}{\D \rvx}} + \log p(\rvz)
\end{align}
The first term on the right (log determinant of the Jacobian) is often intractable, however, previous works on normalizing flows have found ways to estimate this efficiently. The second term, $\log p(\rvz)$, is computed as the log probability of $\rvz$ under a known noise distribution, typically the standard Gaussian $\mathcal{N}(\mathbf{0}, \mathbf{I})$.


\subsection{Wavelet Flow~\citep{yu2020waveletflow}}

WaveletFlow splits an image using the Discrete Wavelet Transformation, and maps the average image at each resolution to noise using a normalizing flow. WaveletFlow builds on the Glow~\citep{kingma2018glow} architecture. It uses an orthogonal transformation, which does not preserve range, and adds a constant term to the log likelihood at each resolution. Best results are obtained when WaveletFlow models with a high parameter count are trained for a long period of time. We aim to fix these issues using our \gls{mrcnf}.

\subsection{Continuous Normalizing Flows}

Continuous Normalizing Flows (CNF)~\citep{chen2018neural, grathwohl2019ffjord, finlay2020rnode} are a variant of normalizing flows that operate in the continuous domain.
A \gls{cnf} creates a geometric flow between the input and target (noise) distributions, by assuming that the state transition is governed by an Ordinary Differential Equation (ODE). It further assumes that the differential function is parameterized by a neural network, this model is called a Neural ODE~\citep{chen2018neural}. Suppose \gls{cnf} $g$ transforms its state $\rvv(t)$ using a Neural ODE, with neural network $f$ defining the differential. Here, $\rvv(t_0)=\rvx$ is, say, an image, and at the final time step $\rvv(t_1)=\rvz$ is a sample from a known noise distribution.
\begin{align*}
\frac{\D \rvv(t)}{\D t} = f(\rvv(t), t)
\implies \rvv(t_1) = g(\rvv(t_0)) = \rvv(t_0) + \int_{t_0}^{t_1} f(\rvv(t), t)\ \D t \numberthis
\label{eq:neural_ode}
\end{align*}
This integration is typically performed by an ODE solver. Since this integration can be run backwards as well to obtain the same $\rvv(t_0)$ from $\rvv(t_1)$, a \gls{cnf} is a reversible model.

\autoref{eq:change_of_variables} can be used to compute the change in log-probability induced by the \gls{cnf}. However, \citet{chen2018neural} and \citet{grathwohl2019ffjord} proposed a more efficient variant in the context of \gls{cnf}s, called the instantaneous change-of-variables formula:
\begin{align*}
\frac{\partial \log p(\rvv(t))}{\partial t} = -\text{Tr}\left(\frac{\partial f}{\partial \rvv(t)}\right) \implies \Delta\log p_{\rvv(t_0) \rightarrow \rvv(t_1)} = -\int_{t_0}^{t_1}\text{Tr}\left(\frac{\partial f}{\partial \rvv(t)} \right) \D t
\numberthis \label{eq:inst}
\end{align*}
Hence, the change in log-probability of the state of the Neural ODE i.e. $\Delta\log p_\rvv$ is expressed as another differential equation. The ODE solver now solves both differential equations \cref{eq:neural_ode} and \cref{eq:inst} by augmenting the original state with the above. Thus, a \gls{cnf} provides both the final state $\rvv(t_1)$ as well as the change in log probability $\Delta\log p_{\rvv(t_0) \rightarrow \rvv(t_1)}$ together.

Prior works~\citep{grathwohl2019ffjord, finlay2020rnode, ghosh2020steer, onken2021otflow, huang2021acc} have trained CNFs as reversible generative models of images, by maximizing the likelihood of the images under the model:
\begin{align}
\rvz = g(\rvx) \quad ;
\qquad
\log p(\rvx) = \Delta\log p_{\rvx \rightarrow \rvz} + \log p(\rvz)
\label{eq:cnf}
\end{align}
where $\rvx$ is an image, $\rvz$ and $\Delta\log p_{\rvx \rightarrow \rvz}$ are computed by the CNF using \cref{eq:neural_ode} and \cref{eq:inst}, and $\log p(\rvz)$ is the likelihood of the computed $\rvz$ under a known noise distribution, typically the standard Gaussian $\mathcal{N}(\mathbf{0}, \mathbf{I})$. Novel images are generated by sampling $\rvz$ from the known noise distribution, and running it through the CNF in reverse.

\section{Our method}
\label{method}
Our method is a reversible generative model of images that builds on top of \gls{cnf}s. We introduce the notion of multiple resolutions in images, and connect the different resolutions in an autoregressive fashion. This helps generate images faster, with better likelihood values at higher resolutions, using only one GPU in all our experiments. We call this model Multi-Resolution Continuous Normalizing Flow (MRCNF).

\subsection{Multi-Resolution image representation}
\label{sec:MRimage}
Multi-resolution representations of images have been explored in computer vision for decades~\citep{burt1981fast, marr2010vision, witkin1987scale, burt1983laplacian, mallat1989theory, lindeberg1990scale}. This implies that much of the content of an image at a resolution is a composition of low-level information captured at coarser resolutions, and some high-level information not present in the coarser images. We take advantage of this property by first decomposing an image in \emph{resolution space} i.e. by expressing it as a series of $S$ images at decreasing resolutions: $\rvx \rightarrow (\rvx_1, \rvx_2, \dots, \rvx_S)$, where $\rvx_1=\rvx$ is the finest image, $\rvx_S$ is the coarsest, and every $\rvx_{s+1}$ is the average image of $\rvx_s$.
This called an image pyramid, or a Gaussian Pyramid if the upsampling-downsampling operations include a Gaussian filter~\citep{burt1981fast, burt1983laplacian, adelson1984pyramid, witkin1987scale, lindeberg1990scale}. In this work, we obtain a coarser image simply by averaging pixels in every $2\x2$ patch, thereby halving the width and height.

However, this representation is redundant since much of the information in $\rvx_1$ is contained in $\rvx_{s>1}$. Instead, we express $\rvx$ as a series of high-level information $\rvy_s$ not present in the immediate coarser images $\rvx_{s+1}$, and a final coarse image $\rvx_S$:
\begin{align}
\rvx \rightarrow (\rvy_1, \rvx_2) \rightarrow (\rvy_1, \rvy_2, \rvx_3) \rightarrow \dots \rightarrow (\rvy_1, \rvy_2, \dots, \rvy_{S-1}, \rvx_S)
\label{eq:multi-resolution}
\end{align}
Our overall method is to map these $S$ terms to $S$ noise samples using $S$ CNFs.


\subsection{Defining the high-level information $\rvy_s$}
\label{subsec:y}

\begin{wrapfigure}{r}{.33\textwidth}
\includegraphics[width=.33\textwidth]{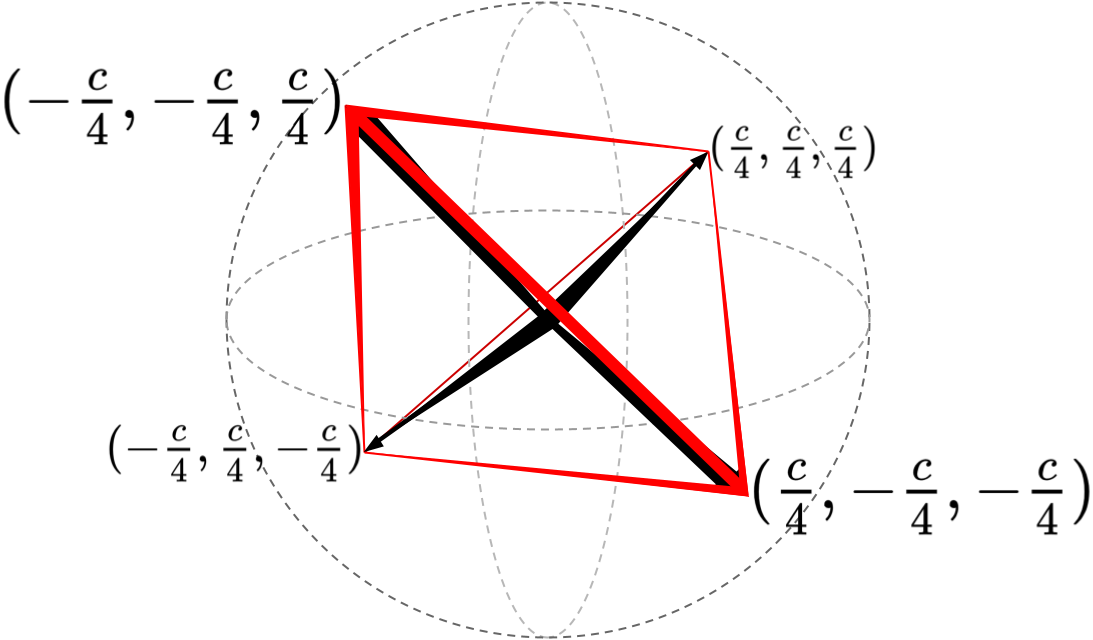}
\caption{\small{Tetrahedron in 3D space with 4 corners}}
\label{fig:Tetrahedron}
\vspace{-1.5em}
\end{wrapfigure}

We choose to design a linear transformation with the following properties: 1) invertible i.e. it should be possible to deterministically obtain $\rvx_s$ from $\rvy_s$ and $\rvx_{s+1}$, and vice versa ;
2) volume preserving i.e. determinant is 1, change in log-likelihood is 0 ; 3) angle preserving ; and 4) range preserving (under the notion of the maximum principle \citep{varga1966discrete}).




Consider the simplest case of 2 resolutions where $\rvx_1$ is a $2\x2$ image with pixel values $x_1, x_2, x_3, x_4$, and $\rvx_2$ is a $1\x1$ image with pixel value $\xbar = \frac{1}{4}(x_1 + x_2 + x_3 + x_4)$. We require three values $(y_1, y_2, y_3) = \rvy_1$ that contain information not present in $\rvx_2$, such that $\rvx_1$ is obtained when $\rvy_1$ and $\rvx_2$ are combined.

This could be viewed as a problem of finding a matrix $\rmM$ such that: $[x_1, x_2, x_3, x_4]^\top = \rmM\ [y_1, y_2, y_3, \xbar]^\top$. We fix the last column of $\rmM$ as $[1, 1, 1, 1]^\top$, since every pixel value in $\rvx_1$ depends on $\barx$. Finding the rest of the parameters can be viewed as requiring four 3D vectors that are spaced such that they do not degenerate the number of dimensions of their span. These can be considered as the four corners of a tetrahedron in 3D space,
under any configuration (rotated in 3D space), and any scaling of the vectors (see \autoref{fig:Tetrahedron}).

Out of the many possibilities for this tetrahedron, we could choose the matrix that performs the Discrete Haar Wavelet Transform~\citep{mallat1989theory, mallat2009wavelet}:
\begin{align}
\label{haar}
\begin{bmatrix}
x_1 \\ x_2 \\ x_3 \\ x_4
\end{bmatrix}
=
\begin{bmatrix}
\ \ \ \frac{1}{2} & \ \ \ \frac{1}{2} & \ \ \ \frac{1}{2} & 1 \\
\ \ \ \frac{1}{2} & -\frac{1}{2} & -\frac{1}{2} & 1 \\
-\frac{1}{2} & \ \ \ \frac{1}{2} & -\frac{1}{2} & 1 \\
-\frac{1}{2} & -\frac{1}{2} & \ \ \ \frac{1}{2} & 1
\end{bmatrix}
\begin{bmatrix}
y_1 \\ y_2 \\ y_3 \\ \xbar
\end{bmatrix}
\iff
\begin{bmatrix}
y_1 \\ y_2 \\ y_3 \\ \xbar
\end{bmatrix}
=
\begin{bmatrix}
\frac{1}{2} & \ \ \ \frac{1}{2} & -\frac{1}{2} & -\frac{1}{2} \\
\frac{1}{2} & -\frac{1}{2} & \ \ \ \frac{1}{2} & -\frac{1}{2} \\
\frac{1}{2} & -\frac{1}{2} & -\frac{1}{2} & \ \ \ \frac{1}{2} \\
\frac{1}{4} & \ \ \ \frac{1}{4} & \ \ \ \frac{1}{4} & \ \ \ \frac{1}{4}
\end{bmatrix}
\begin{bmatrix}
x_1 \\ x_2 \\ x_3 \\ x_4
\end{bmatrix}
\end{align}
However, this has $\log \Abs{\det (\rmM^{-1})} = \log(1/2)$ (\cref{haar}), and is therefore not volume preserving. Other simple scaling of \cref{haar} has been used in the past, for example multiplying the last row of \cref{haar} by 2, yielding an orthogonal transformation, such as in WaveletFlow~\citep{yu2020waveletflow}. However, this transformation neither preserves the volume i.e. the log determinant is not 0, nor the maximum i.e. the range of $\rvx_s$ changes.  

We wish to find a transformation $\rmM$ where: one of the results is the average of the inputs, $\bar x$; it is unit determinant; the columns are orthogonal; and it preserves the range of $\bar x$.
Fortunately such a matrix exists -- although we have not seen it discussed in prior literature. It can be seen as a variant of the Discrete Haar Wavelet Transformation matrix that is unimodular, i.e. has a determinant of $1$ (and is therefore volume preserving), while also preserving the range of the images for the input and its average:
\begin{align}
\label{Mdefn}
\begin{bmatrix}
x_1 \\ x_2 \\ x_3 \\ x_4
\end{bmatrix}
=
\frac{1}{a}
\begin{bmatrix}
\ \ \ c & \ \ \ c & \ \ \ c & a \\
\ \ \ c & -c & -c & a \\
-c & \ \ \ c & -c & a \\
-c & -c & \ \ \ c & a
\end{bmatrix}
\begin{bmatrix}
y_1 \\ y_2 \\ y_3 \\ \xbar
\end{bmatrix}
\hspace{-.5em}
\iff
\hspace{-.5em}
\begin{bmatrix}
y_1 \\ y_2 \\ y_3 \\ \xbar
\end{bmatrix}
=
\begin{bmatrix}
c^{-1} & \ \ \ c^{-1} & -c^{-1} & -c^{-1}\\
c^{-1} & -c^{-1} & \ \ \ c^{-1} & -c^{-1}\\
c^{-1} & -c^{-1} & -c^{-1} & \ \ \ c^{-1}\\
a^{-1} & \ \ \ a^{-1} & \ \ \ a^{-1} & \ \ \ a^{-1}
\end{bmatrix}
\begin{bmatrix}
x_1 \\ x_2 \\ x_3 \\ x_4
\end{bmatrix}
\end{align}
where $c = 2^{2/3}$, $a=4$. Hence, $\log \Abs{\det (\rmM^{-1})} = \log(1)=0$.
This can be scaled up to larger spatial regions by performing the same calculation for each $2\x2$ patch. Let $M$ be the function that uses matrix $\rmM$ from above and combines every pixel in $\rvx_{s+1}$ with the three corresponding pixels in $\rvy_s$ to make the $2\x2$ patch at that location in $\rvx_s$ using \cref{Mdefn}:
\begin{align}
\rvx_s = M(\rvy_s, \rvx_{s+1}) \iff \rvy_s, \rvx_{s+1} = M^{-1}(\rvx_s)
\label{eq:M}
\end{align}
\autoref{eq:change_of_variables} can be used to compute the change in log likelihood from this transformation $\rvx_s \rightarrow (\rvy_s, \rvx_{s+1})$:
\begin{align}
&\log p(\rvx_s) = \Delta \log p_{\rvx_s \rightarrow (\rvy_s, \rvx_{s+1})} + \log p(\rvy_s, \rvx_{s+1}) = \log \Abs{\det (M^{-1})} + \log p(\rvy_s, \rvx_{s+1})
\label{eq:multires_logp}
\end{align}
where $\log \Abs{\det (M^{-1})} = \text{dims}(\rvx_{s+1})\log (1/2)$ in the case of \cref{haar}, where ``dims'' is the number of pixels times the number of channels (typically 3) in the image, and $\log \Abs{\det (M^{-1})} = 0$ for \cref{Mdefn}. 

\subsection{Multi-Resolution Continuous Normalizing Flows}

Using the multi-resolution image representation in \cref{eq:multi-resolution}, we characterize the conditional distribution over the additional degrees of freedom ($\rvy_s$) required to generate a higher resolution image ($\rvx_s$) that is consistent with the average ($\rvx_{s+1}$) over the equivalent pixel space. At each resolution $s$, we use a CNF to reversibly map between $\rvy_s$ (or $\rvx_S$ when $s\mkern1mu{=}\mkern1mu S$) and a sample $\rvz_s$ from a known noise distribution. For generation, $\rvy_s$ only adds 3 degrees of freedom to $\rvx_{s+1}$, which contain information missing in $\rvx_{s+1}$, but conditional on it.

This framework ensures that one coarse image could generate several potential fine images, but these fine images have the same coarse image as their average. This fact is preserved across resolutions.
Note that the 3 additional pixels in $\rvy_s$ per pixel in $\rvx_{s+1}$ are generated conditioned on the entire coarser image $\rvx_{s+1}$, thus maintaining consistency using the full context.

In principle, any generative model could be used to map between the multi-resolution image and noise. Normalizing flows are good candidates for this as they are probabilistic generative models that perform exact likelihood estimates, and can be run in reverse to generate novel data from the model's distribution. This allows model comparison and measurement of generalization to unseen data. We choose to use the CNF variant of normalizing flows at each resolution. CNFs have recently been shown to be effective in modeling image distributions using a fraction of the number of parameters typically used in normalizing flows (and non flow-based approaches), and their underlying framework of Neural ODEs have been shown to be more robust than convolutional layers~\citep{yan2020robustness}.


\textbf{Training}: We train an MRCNF by maximizing the average log-likelihood of the images in the training dataset under the model.
The log probability of each image $\log p(\rvx)$ can be estimated recursively from \cref{eq:multires_logp} as:
\begin{align*}
\log p(\rvx)
&= \Delta\log p_{\rvx_1 \rightarrow (\rvy_1, \rvx_2)} + \log p(\rvy_1, \rvx_2) = \Delta\log p_{\rvx_1 \rightarrow (\rvy_1, \rvx_2)} + \log p(\rvy_1 \mid \rvx_2) + \log p(\rvx_2)\\
&= \sum_{s=1}^{S-1}\left(\Delta\log p_{\rvx_s \rightarrow (\rvy_s, \rvx_{s+1})} + \log p(\rvy_s \mid \rvx_{s+1})\right) + \log p(\rvx_S) \numberthis
\label{eq:loglikelihood_recurse}
\end{align*}
where $\Delta\log p_{\rvx_s \rightarrow (\rvy_s, \rvx_{s+1})}$ is given by \cref{eq:multires_logp}, $\log p(\rvy_s \mid \rvx_{s+1})$ and $\log p(\rvx_S)$ are given by \cref{eq:cnf}:
\begin{align}
\label{eq:train_cnf}
\rvz_s = g_s(\rvy_s \mid \rvx_{s+1}) \quad &;
\qquad
\log p(\rvy_s \mid \rvx_{s+1}) = \Delta\log p_{(\rvy_s \rightarrow \rvz_s) \mid \rvx_{s+1}} + \log p(\rvz_s)\\
\rvz_S = g_S(\rvx_S) \quad &;
\qquad
\log p(\rvx_S) = \Delta\log p_{\rvx_S \rightarrow \rvz_S} + \log p(\rvz_S)
\end{align}
The coarsest resolution $S$ can be chosen such that the last CNF operates on the image distribution at a small enough resolution that is easy to model unconditionally. All other CNFs are conditioned on the immediate coarser image. The conditioning itself is achieved by concatenating the input image of the CNF with the coarser image. This model could be seen as a stack of CNFs connected in an autoregressive fashion.

Typically, likelihood-based generative models are compared using the metric of bits-per-dimension (BPD), i.e. the negative log likelihood per pixel in the image. Hence, we train our MRCNF to minimize the average BPD of the images in the training dataset, computed using \cref{eq:bpd}:
\begin{align}
\text{BPD}(\rvx) = -\log p(\rvx)/\text{dims}(\rvx)
\label{eq:bpd}
\end{align}

We use FFJORD~\citep{grathwohl2019ffjord} as the baseline model for our \gls{cnf}s. In addition, we use to two regularization terms introduced by RNODE~\citep{finlay2020rnode} to speed up the training of FFJORD models by stabilizing the learnt dynamics: the kinetic energy of the flow $\mathcal{K}(\theta)$, and the Jacobian norm $\mathcal{B}(\theta)$:
\begin{align}
\mathcal{K}(\theta) = \int_{t_0}^{t_1} \| f(\rvv(t), t, \theta) \|_2^2\ \D t ;
\qquad \mathcal{B}(\theta) = \int_{t_0}^{t_1} \| \epsilon^\top \nabla_z f(\rvv(t), t, \theta) \|_2^2\ \D t, \quad \epsilon \sim \mathcal{N}(0, I)
\end{align}

\textbf{Parallel training:} Note that although the final log likelihood $\log p(\rvx)$ involves sequentially summing over values returned by all $S$ CNFs, the log likelihood term of each CNF is independent of the others. Conditioning is done using ground truth images. Hence, each CNF can be trained independently, in parallel.

\textbf{Generation}: Given an $S$-resolution model, we first sample $\rvz_s, s={1,\dots,S}$ from the latent noise distributions. The \gls{cnf} $g_s$ at resolution $s$ transforms the noise sample $\rvz_s$ to high-level information $\rvy_s$ conditioned on the immediate coarse image $\rvx_{s+1}$ (except $g_S$ which is unconditioned). $\rvy_s$ and $\rvx_{s+1}$ are then combined to form $\rvx_s$ using $M$ from \cref{Mdefn}. This process is repeated progressively from coarser to finer resolutions, until the finest resolution image $\rvx_1$ is computed (see \autoref{fig:mrcnf_2d_gen}). It is to be noted that the generated image at one resolution is used to condition the CNF at the finer resolution.
\begin{align*}
\begin{cases}
\rvx_S = g_S^{-1}(\rvz_S) \qquad\qquad &s=S\\
\rvy_s = g_s^{-1}(\rvz_s \mid \rvx_{s+1}) ; \quad
\rvx_s = M(\rvy_s, \rvx_{s+1}) \qquad\qquad
&s = S\text{-}1 \rightarrow 1
\end{cases}
\numberthis
\label{eq:gen_cnf}
\end{align*}



\subsection{Multi-Resolution Noise}
\label{subsec:mrnoise}

We further decompose the noise image as well into its respective coarser components. This means that ultimately we use only one noise image at the finest level, but it is decomposed into multiple resolutions using \cref{Mdefn}. $\rvx_{s+1}$ is mapped to noise of a quarter variance, while $\rvy_s$ is mapped to noise of $c$-factored variance (see \cref{fig:mrcnf_2d_gen}). Although this is optional, it preserves interpretation between the single- and multi-resolution models.

\section{Related work}
\label{section:related}
Multi-resolution approaches already serve as a key component of state-of-the-art GAN~\citep{denton2015lapgan, karras2017progressive, karnewar2020msg} and VAE~\citep{razavi2019generating, vahdat2020nvae} based deep generative models.
Deconvolutional CNNs~\citep{long2015fully, radford2015unsupervised} use upsampling layers to generate images more effectively. Modern state-of-the-art generative models have also injected noise at different levels to improve sample quality \citep{brock2019biggan, karras2020analyzing,vahdat2020nvae}. 

Several prior works on normalizing flows~\citep{kingma2018glow, hoogeboom2019emerging, hoogeboom2019idf, song2019mintnet, ma2019macow, durkan2019spline, chen2020vflow, ho2019flow++, lee2020nanoflow, yu2020waveletflow} build on RealNVP~\citep{dinh2016density}. Although they achieve great results in terms of \gls{bpd} and image quality, they nonetheless report results from significantly higher number of parameters (some with 100x!), and several times GPU hours of training.

STEER~\citep{ghosh2020steer} introduced temporal regularization to CNFs by making the final time of integration stochastic. However, we found that this increased training time without significant BPD improvement.

\textbf{Comparison to WaveletFlow}: We emphasize that there are important and crucial differences between our \gls{mrcnf} and WaveletFlow. We generalize the notion of a multi-resolution image representation (\cref{subsec:y}), and show that Wavelets are one case of this general formulation. WaveletFlow builds on the Glow~\citep{kingma2018glow} architecture, while ours builds on \gls{cnf}s~\citep{grathwohl2019ffjord, finlay2020rnode}. We also make use of the notion of multi-resolution decomposition of the noise, which is optional, but is not taken into account by WaveletFlow. WaveletFlow uses an orthogonal transformation which does not preserve range ; our MRCNF uses \cref{Mdefn} which is volume-preserving and range-preserving. Finally, WaveletFlow applies special sampling techniques to obtain better samples from its model. We have so far not used such techniques for generation, but we believe they can potentially help our models as well. By making these important changes, we fix many of the previously discussed issues with WaveletFlow. For a more detailed ablation study, please check \autoref{subsec:waveletflow_ablation}.

\textbf{``Multiple scales'' in prior normalizing flows}: Normalizing flows~\citep{dinh2016density, kingma2018glow, grathwohl2019ffjord} try to be ``multi-scale'' by transforming the input in a smart way (squeezing operation) such that the width of the features progressively reduces in the direction of image to noise, while maintaining the total dimensions. This happens while operating at a \textit{single resolution}. In contrast, our model stacks normalizing flows at \textit{multiple resolutions} in an autoregressive fashion by conditioning on the images at coarser resolutions.


\section{Experimental results}
\label{experiments}
\begin{table*}[!htb]
    \vspace{-1em}
    \small
    \centering
    {\def\arraystretch{1.1}
    \setlength{\tabcolsep}{.1em}   \caption{\normalsize{Bits-per-dimension (lower is better) of images in the corresponding evaluation sets for CIFAR10, ImageNet $32\x32$, and ImageNet $64\x64$. We also report the number of parameters in the models, and the time taken to train (in GPU hours). All our models were trained on only one GPU.\\
    $^{ }$ Blank spaces indicate unreported values.
    \quad
    $^\ddagger$As reported in~\cite{ghosh2020steer}.
    \quad 
    $^\mathsection$Re-implemented by us.
    \quad
    `x': Fails to train.
    \quad
    $^*$RNODE~\cite{finlay2020rnode} used 4 GPUs to train on ImageNet64.}}
    \begin{tabular}{|l|crc|rrc|crc|}
        \hline
        & \multicolumn{3}{c|}{\textbf{\textsc{CIFAR10}}} & \multicolumn{3}{c|}{\textbf{\textsc{ImageNet32}}} & \multicolumn{3}{c|}{\textbf{\textsc{ImageNet64}}} \\
        & \textsc{BPD} & \textsc{Param} & \textsc{Time}
        & \textsc{BPD} & \textsc{Param} & \textsc{Time}
        & \textsc{BPD} & \textsc{Param} & \textsc{Time} \\
        \hline\hline
        \multicolumn{10}{|l|}{\textbf{Non Flow-based Prior Work}} \\
        \hline
        Gated PixelCNN
        ~\citep{van2016conditional}
            & 3.03 &  & 
            & 3.83 &  & 60
            & 3.57 &  & 60 \\[-2pt]
        SPN
        ~\citep{menick2019generating}
            &  &  & 
            & 3.85 & \footnotesize{150.0M} &  
            & 3.53 & \footnotesize{150.0M} &   \\[-2pt]
        Sparse Transformer
        ~\citep{child2019generating}
            & 2.80 & \footnotesize{59.0M} & 
            &  &  &  
            & 3.44 & \footnotesize{152.0M} & \footnotesize{7days}  \\[-2pt]
        NVAE
        ~\citep{vahdat2020nvae}
            & 2.91 &  & 55
            & 3.92 &  & 70 
            &  &  &   \\[-2pt]
        DistAug
        ~\citep{jun2020distaug}
            & 2.56 & \footnotesize{152.0M} &
            &  &  & 
            & 3.42 & \footnotesize{152.0M} &  \\
        \hline
        \hline
        \multicolumn{10}{|l|}{\textbf{Flow-based Prior Work}} \\
        \hline
        RealNVP
        ~\citep{dinh2016density}
            & 3.49 & & 
            & 4.28 & \footnotesize{46.0M} & 
            & 3.98 & \footnotesize{96.0M} & \\[-2pt]
        Glow
        ~\citep{kingma2018glow}
            & 3.35 & \footnotesize{44.0M} &
            & 4.09 & \footnotesize{66.1M} & 
            & 3.81 & \footnotesize{111.1M} & \\[-2pt]
        MaCow
        ~\citep{ma2019macow}
            & 3.16 & \footnotesize{43.5M} &
            &  &  & 
            & 3.69 & \footnotesize{122.5M} & \\[-2pt]
        Flow++
        ~\citep{ho2019flow++}
            & 3.08 & \footnotesize{31.4M} &
            & 3.86  & \footnotesize{169.0M} &
            & 3.69 & \footnotesize{73.5M} & \\[-2pt]
        Wavelet Flow
        ~\citep{yu2020waveletflow}
            &  &  & 
            & 4.08 & \footnotesize{64.0M} & 
            & 3.78 & \footnotesize{96.0M} & 822  \\[-2pt]
        DenseFlow
        ~\citep{grcic2021denseflow}
            & 2.98 &  & 250
            & 3.63 &  & 310
            & 3.35 &  & 224  \\
        \hline
        \hline
        \multicolumn{10}{|l|}{\textbf{1-Resolution Continuous Normalizing Flow}}\\
        \hline
        FFJORD
        ~\citep{grathwohl2019ffjord}
            & 3.40 & 0.9M & \footnotesize{$\geq$5days}
            & $^{\ddagger}3.96$ & $^{\ddagger}$\footnotesize{2.0M} & $^{\ddagger}$\footnotesize{$>$5days}
            & x &  & x \\[-0pt]
        RNODE
        ~\citep{finlay2020rnode}
            & 3.38 & 1.4M & 31.8
            & $^{\ddagger}2.36$ & \footnotesize{2.0M} & $^{\ddagger}30.1$
            & $^*3.83$ & \footnotesize{2.0M} & $^*256.4$ \\[-3pt]
            &  &  & 
            & $^{\mathsection}3.49$ & $^\mathsection$\footnotesize{1.6M} & $^{\mathsection}40.4$
            & & &  \\
        FFJORD + STEER
        ~\citep{ghosh2020steer}
            & 3.40 & \footnotesize{1.4M} & 86.3
            & 3.84 & \footnotesize{2.0M} & \footnotesize{$>$5days}
            & & & \\
        RNODE + STEER
        ~\citep{ghosh2020steer}
            & 3.397 & \footnotesize{1.4M} & 22.2
            & 2.35 & \footnotesize{2.0M} & 24.9
            & & & \\[-3pt]
            & & & 
            & $^\mathsection3.49$ & $^\mathsection$\footnotesize{1.6M} & $^\mathsection30.1$
            & & & \\
        \hline
        \hline
        \multicolumn{10}{|l|}{\textbf{\textsc{(Ours)} Multi-Resolution Continuous Normalizing Flow (MRCNF)}}\\
        \hline
        2-resolution MRCNF
            & 3.65 & \footnotesize{1.3M} & 19.8
            & 3.77 & \footnotesize{1.3M} & 18.2
            & 3.44& \footnotesize{2.0M} & 42.3 \\
        2-resolution MRCNF
            & 3.54 & \footnotesize{3.3M} & 36.5
            & 3.78 & \footnotesize{6.7M} & 18.0
            & x & \footnotesize{6.7M} & x \\
        3-resolution MRCNF
            & 3.79 & \footnotesize{1.5M} & 17.4
            & 3.97 & \footnotesize{1.5M} & 13.8
            & 3.55 & \footnotesize{2.0M} & 35.4 \\
        3-resolution MRCNF
            & 3.60 & \footnotesize{5.1M} & 38.3
            & 3.93 & \footnotesize{10.2M} & 41.2
            & x & \footnotesize{7.6M} & x\\
        \hline
    \end{tabular}
    \label{tab:msflow}
    }
\end{table*}

We train \gls{msflow-image} models on  the CIFAR10~\citep{krizhevsky2009learning} dataset at finest resolution of 32x32, and the ImageNet~\citep{deng2009imagenet} dataset at 32x32, 64x64, 128x128. We build on top of the code provided in \citet{finlay2020rnode}\footnote{https://github.com/cfinlay/ffjord-rnode}. In all cases, we train using \textit{only one} NVIDIA RTX 20280 Ti GPU with 11GB.

In \autoref{tab:msflow}, we compare our results with prior work in terms of (lower is better in all cases) the \gls{bpd} of the images of the test datasets under the trained models, the number of parameters used by the model, and the number of GPU hours taken to train.
The most relevant models for comparison are the 1-resolution FFJORD~\citep{grathwohl2019ffjord} models, and their regularized version RNODE~\citep{finlay2020rnode}, since our model directly converts their architecture into multi-resolution. Other relevant comparisons are previous flow-based methods~\citep{dinh2016density, kingma2018glow, song2019mintnet, ho2019flow++, yu2020waveletflow}, however their core architecture (RealNVP~\citep{dinh2016density}) is quite different from FFJORD.

\textbf{BPD}: At lower resolution spaces, we achieve comparable \gls{bpd}s in lesser time with far fewer parameters than previous normalizing flows (and non flow-based approaches). However, the power of the multi-resolution formulation is more evident at higher resolutions: we achieve better BPD for ImageNet64 with significantly fewer parameters and lower time using only one GPU.


It is to be noted that we were not able to reproduce the same BPD as provided by STEER~\citep{ghosh2020steer}, we report the results of our re-implementation. A more complete table can be found in the appendix.

\textbf{Train time}: All our experiments used only one GPU, and took significantly less time to train than 1-resolution CNFs, and all prior works including flow-based and non-flow-based models. 
Since all the \gls{cnf}s can be trained in parallel, the actual training time in practice could be much lower than reported.

\begin{wrapfigure}{r}{0.6\linewidth}
    \centering
    \includegraphics[width=.6\textwidth]{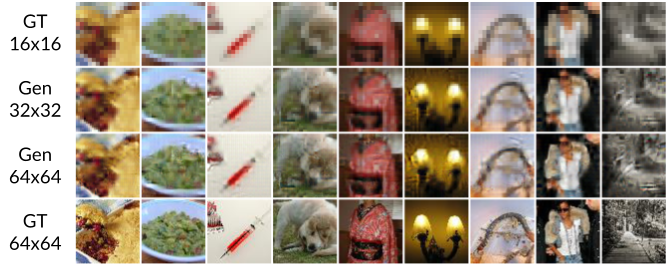}
    \caption{ImageNet: Example of super-resolving from ground truth $16\x16$ to $64\x64$. Top: ground truth, middle: generated, bottom: ground truth. }
    \label{fig:super-res-images}
    \vspace{-1em}
\end{wrapfigure}

\textbf{Super-resolution}: Our formulation also allows for super-resolution of images (\autoref{fig:super-res-images}) free of cost since our framework is autoregressive in resolution.


\begin{table}
\vspace{-1.5em}
    \small
    \centering
    \caption{\small{Metrics for unconditional ImageNet128 generation.
    }}
    \vspace{-0.5em}
    \begin{tabular}{|l@{\hskip 4pt}|@{\hskip 4pt}c@{\hskip 8pt}c@{\hskip 8pt}c@{\hskip 4pt}|}
        \hline
        \textbf{\textsc{ImageNet128}}
        & \textsc{BPD} & \textsc{Param} & \textsc{Time}\Tstrut
        \\\hline
        Parallel Multiscale~\cite{reed2017parallel} & 3.55 & &\Tstrut
        \\
        SPN~\cite{menick2019generating} & 3.08 & \footnotesize{250M} &
        \\
        \hline
        \textbf{\textsc{(Ours)} 4-resolution MRCNF} & 3.31 \scriptsize{$\pm$0.69} & \footnotesize{2.74M} & 58.59\Tstrut
        \\\hline
    \end{tabular}
    \label{tab:imagenet128}
\end{table}

\textbf{Progressive training}: We trained an MRCNF model on ImageNet128 by training only the finest resolution ($128\x128$) conditioned on the immediate coarser ($64\x64$) images, and attached it to a 3-resolution model trained on ImageNet64. The resultant 4-resolution ImageNet128 model gives a \gls{bpd} of 3.31 (\autoref{tab:imagenet128}) with just 2.74M parameters in $\approx$60 GPU hours.



\subsection{Ablation study}
\label{subsec:waveletflow_ablation}

Our MRCNF method differs from WaveletFlow in three respects: 1. we use CNFs, 2. we use \cref{Mdefn} instead of \cref{haar} as used by WaveletFlow, 3. we use multi-resolution noise. We check the individual effects of these changes in an ablation study in Table \ref{tab:ablation}, and conclude that:

\begin{table*}[!htb]
    \small
    \centering
    {\def\arraystretch{1.1}
    \setlength{\tabcolsep}{.4em}
    \caption{\normalsize{Ablation study across using Wavelet in \cref{haar}, and multi-resolution noise formulation in \ref{subsec:mrnoise}.}}
    \vspace{-.5em}
    \begin{tabular}{|l|crr|crr|}
        \hline
        & \multicolumn{3}{c|}{\textbf{\textsc{CIFAR10}}} & \multicolumn{3}{c|}{\textbf{\textsc{ImageNet64}}} \\
        & \textsc{BPD} & \textsc{Param} & \textsc{Time}
        & \textsc{BPD} & \textsc{Param} & \textsc{Time} \\
        \hline
        WaveletFlow \citep{yu2020waveletflow}
            &  &  & 
            & 3.78 & \footnotesize{98.0M} & 822.00 \\
        1-resolution CNF (RNODE) \citep{finlay2020rnode}
            & 3.38 & \footnotesize{1.4M} & 31.84
            & 3.83 & \footnotesize{2.0M} & 256.40 \\
        \hline\hline
        \multicolumn{7}{|l|}{\textbf{2-resolution}}\\
        \hline
        \cref{haar} WaveletFlow, but with CNFs, w/o multi-res noise
            & 3.68 & \footnotesize{1.3M} & 27.25
            & x & \footnotesize{2.0M} & x \\
        \cref{haar} WaveletFlow, but with CNFs, w/ multi-res noise
            & 3.69 & \footnotesize{1.3M} & 25.88
            & x & \footnotesize{2.0M} & x \\
        \cref{Mdefn} MRCNF w/o multi-res noise
            & 3.66 & \footnotesize{1.3M} & 19.79
            & 3.48 & \footnotesize{2.0M} & 42.33 \\
        \cref{Mdefn} MRCNF w/ multi-res noise \textbf{(Ours)}
            & 3.65 & \footnotesize{1.3M} & 19.69
            & 3.44 & \footnotesize{2.0M} & 42.30 \\
        \hline\hline
        \multicolumn{7}{|l|}{\textbf{3-resolution}}\\
        \hline
        \cref{haar} WaveletFlow, but with CNFs, w/o multi-res noise
            & 3.82 & \footnotesize{1.5M} & 22.99
            & 3.62 & \footnotesize{2.0M} & 43.37 \\
        \cref{haar} WaveletFlow, but with CNFs, w/ multi-res noise
            & 3.82 & \footnotesize{1.5M} & 25.28
            & 3.62 & \footnotesize{2.0M} & 44.21 \\
        \cref{Mdefn} MRCNF w/o multi-res noise
            & 3.79 & \footnotesize{1.5M} & 17.25
            & 3.57 & \footnotesize{2.0M} & 35.42 \\
        \cref{Mdefn} MRCNF w/ multi-res noise \textbf{(Ours)}
            & 3.79 & \footnotesize{1.5M} & 17.44
            & 3.55 & \footnotesize{2.0M} & 35.39 \\
        \hline
    \end{tabular}
    \label{tab:ablation}
    }
    \vspace{-1em}
\end{table*}

\begin{enumerate}
\item Simply replacing the normalizing flows in WaveletFlow with CNFs does not produce the best results. It does improve the BPD and training time compared to WaveletFlow.
\item Using our unimodular transformation in \cref{Mdefn} instead of the original Wavelet Transformation of \cref{haar} not only improves the BPD, it also consistently decreases training time.
\item As expected, the use of multi-resolution noise does not have a critical impact on either BPD or training time. We use it anyway so as to retain interpretation with 1-resolution models.
\end{enumerate}
Thus, our MRCNF model is not a trivial replacement of normalizing flows with CNFs in WaveletFlow.
We generalize the notion of multi-resolution image representation, in which the Discrete Wavelet Transform is one of many possibilities. We then derived a unimodular transformation that adds no change in likelihood.

\clearpage

\vspace{-1em}
\section{Examining Out-of-Distribution behaviour}
\label{ood}

\begin{wrapfigure}{r}{.55\textwidth}
\vspace{-1.5em}
\centering
\includegraphics[width=\linewidth]{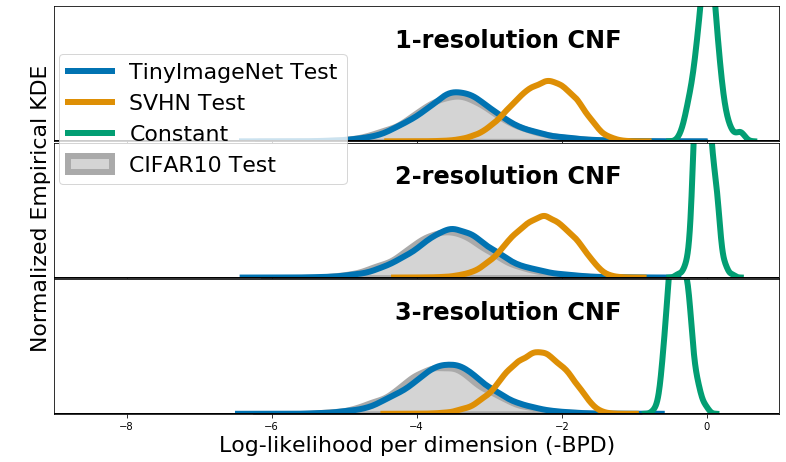}
\caption{\small{Histogram of log likelihood per dimension of out-of-distribution datasets (TinyImageNet, SVHN, Constant) under (MR)CNF models trained on CIFAR10. As with other likelihood-based generative models such as Glow \& PixelCNN, OoD datasets have higher likelihood under (MR)CNFs.}}
\label{fig:ood_bpd}
\vspace{-1em}
\end{wrapfigure}

The derivation of likelihood-based models suggests that the density of an image under the model is an effective measure of its likelihood of being in distribution. However, recent works~\citep{theis2016note, nalisnick2018deep, serra2019input, nalisnick2019detecting} have pointed out that it is possible that images drawn from other distributions have higher model likelihood. Examples have been shown where normalizing flow models (such as Glow) trained on CIFAR10 images assign higher likelihood to SVHN~\citep{Netzer2011svhn} images. This could have serious implications on the practical applicability of these models. Some also note that likelihood-based models do not generate images with good sample quality as they avoid assigning small probability to \gls{ood} data points, hence using model likelihood (-\gls{bpd}) for detecting \gls{ood} data is not effective.

We conduct the same experiments with (MR)CNFs, and find that similar conclusions can be drawn. \autoref{fig:ood_bpd} plots the histogram of log likelihood per dimension (-\gls{bpd}) of \gls{ood} images (SVHN, TinyImageNet) under \gls{msflow-image} models trained on CIFAR10. It can be observed that the likelihood of the \gls{ood} SVHN is higher than CIFAR10 for MRCNF, similar to the findings for Glow, PixelCNN, VAE in earlier works~\citep{nalisnick2018deep, choi2018waic, serra2019input, nalisnick2019detecting, kirichenko2020normalizing}.

One possible explanation put forward by \citet{nalisnick2019detecting} is that ``typical'' images are less ``likely'' than constant images, which is a consequence of the distribution of a Gaussian in high dimensions. Indeed, as our \autoref{fig:ood_bpd} shows, constant images have the highest likelihood under MRCNFs, while randomly generated (uniformly distributed) pixels have the least likelihood (not shown in figure due to space constraints).

\citet{choi2018waic, nalisnick2019detecting} suggest using ``typicality'' as a better measure of \gls{ood}.
However, \citet{serra2019input} observe that the complexity of an image plays a significant role in the training of likelihood-based generative models. They propose a new metric $S$ as an out-of-distribution detector:
\begin{align}
    S(\rvx) = \text{bpd}(\rvx) - L(\rvx)
\end{align}
where $L(\rvx)$ is the complexity of an image $\rvx$ measured as the length of the best compressed version of $\rvx$ (we use FLIF~\citep{sneyers2016flif} following \citet{serra2019input}) normalized by the number of dimensions.

\begin{wraptable}{r}{0.4\linewidth}
\small
\setlength{\tabcolsep}{3pt}
\centering
\caption{\small{auROC for \gls{ood} detection using -bpd and $S$\cite{serra2019input}, for models trained on CIFAR10.}}
\vspace{-.5em}
\begin{tabular}{lcccc}
\textbf{CIFAR10} & \multicolumn{2}{c}{SVHN} & \multicolumn{2}{c}{TIN}
\\
\cline{2-5}
(trained) & -bpd & S
& -bpd & S \\
\hline
\multicolumn{1}{l|}{Glow} & 0.08 & 0.95 & 0.66 & 0.72
\\[0pt]
\multicolumn{1}{l|}{1-res CNF} & 0.07 & 0.16 & 0.48 & 0.60
\\[0pt]
\multicolumn{1}{l|}{2-res MRCNF} & 0.06 & 0.25 & 0.46 & 0.66
\\[0pt]
\multicolumn{1}{l|}{3-res MRCNF} & 0.05 & 0.25 & 0.46 & 0.66
\\
\hline
\end{tabular}
\label{tab:auROC}
\vspace{-1em}
\end{wraptable}

We perform a similar analysis as \citet{serra2019input} to test how $S$ compares with -$\text{bpd}$ for \gls{ood} detection. For different \gls{msflow-image} models trained on CIFAR10, we compute the area under the receiver operating characteristic curve (auROC) using -bpd and $S$ as standard evaluation for the \gls{ood} detection task~\citep{hendrycks2019deep, serra2019input}.

\autoref{tab:auROC} shows that $S$ does perform better than -$\text{bpd}$ in the case of (MR)CNFs, similar to the findings in \citet{serra2019input} for Glow and PixelCNN++. It seems that SVHN is easier to detect as OoD for Glow than MRCNFs. However, OoD detection performance is about the same for TinyImageNet. We also observe that MRCNFs are better at OoD than CNFs. 

Other \gls{ood} methods \citep{hendrycks2017baseline, liang2018enhancing, lee2018simple, sabeti2019data, host2019data, hendrycks2019deep} are not suitable in our case, as identified in \citet{serra2019input}.

\clearpage
\subsection{Shuffled in-distribution images}

\begin{figure*}[thb]
    \centering
    \begin{subfigure}[b]{0.39\textwidth}
     \centering
     \includegraphics[width=\textwidth]{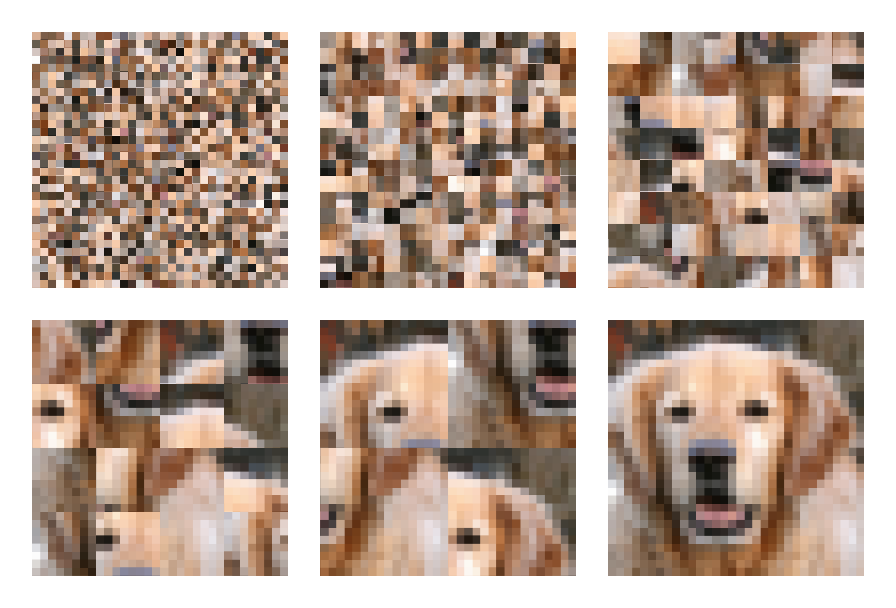}
     \caption{}
     \label{fig:1sc}
    \end{subfigure}
    \hfill
    \begin{subfigure}[b]{0.6\textwidth}
     \centering
     \includegraphics[width=\textwidth]{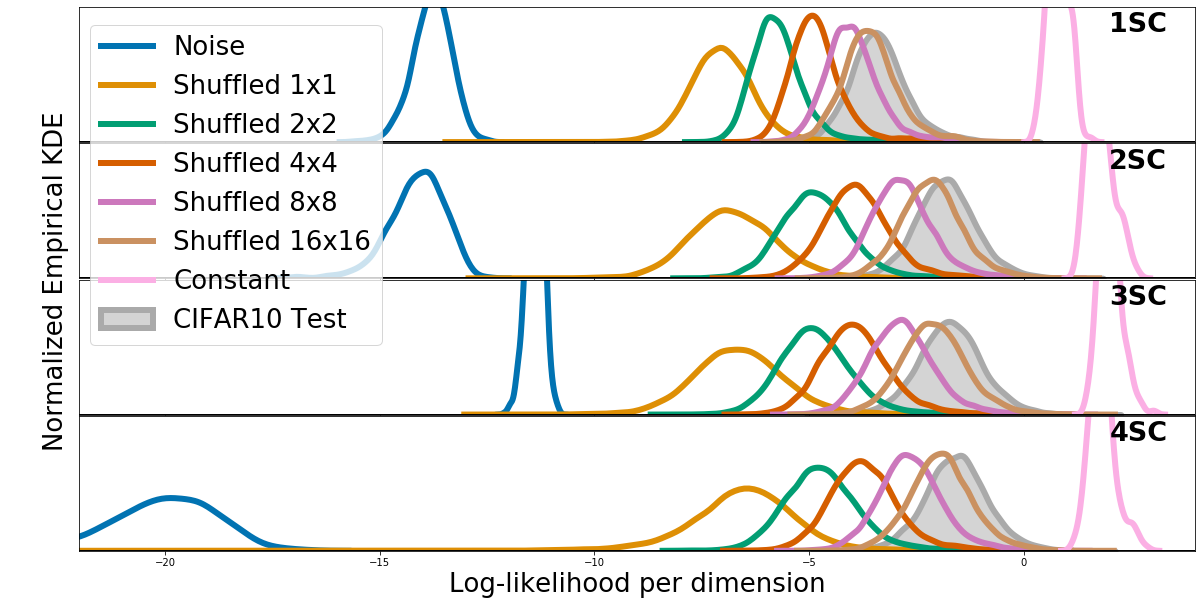}
     \caption{}
     \label{fig:2sc}
    \end{subfigure}
    \caption{\small{(a) Example of shuffling different-sized patches of a $32\x32$ image: (left to right, top to bottom) $1\x1$, $2\x2$, $4\x4$, $8\x8$, $16\x16$, $32\x32$ (unshuffled)} \small{(b) Bits-per-dim vs Epoch at each resolution for different \gls{msflow-image} models trained on CIFAR10.}}
    \label{fig:ood_shuffle}
    \vspace{-1.5em}
\end{figure*}

\citet{kirichenko2020normalizing} conclude that normalizing flows do not represent images based on their semantic contents, but rather directly encode their visual appearance. We verify this for continuous normalizing flows by estimating the density of in-distribution test images, but with patches of pixels randomly shuffled. \autoref{fig:ood_shuffle} (a) shows an example of images of shuffled patches of varying size, \autoref{fig:ood_shuffle} (b) shows the graph of the their log-likelihoods.

That shuffling pixel patches would render the image semantically meaningless is reflected in the \gls{fid} between CIFAR10-Train and these sets of shuffled images --- 1x1: 340.42, 2x2: 299.99, 4x4: 235.22, 8x8: 101.36, 16x16: 33.06, 32x32 (i.e. CIFAR10-Test): 3.15. However, we see that images with large pixel patches shuffled are quite close in likelihood to the unshuffled images, suggesting that since their visual content has not changed much they are almost as likely as unshuffled images under MRCNFs.

\section{Conclusion}
\label{conclusion}
We have presented a Multi-Resolution approach to Continuous Normalizing Flows (MRCNF). MRCNF models achieve comparable or better performance in significantly less training time, training on a single GPU, with a fraction of the number of parameters of other competitive models. Although the likelihood values for $32\x32$ resolution datasets such as CIFAR10 and ImageNet32 do not improve over the baseline, ImageNet64 and above see a marked improvement.
The performance is better for higher resolutions, as seen in the case of ImageNet128. We also conducted an ablation study to note the effects of each change we introduced in the formulation.

In addition, we show that (Multi-Resolution) Continuous Normalizing Flows have similar out-of-distribution properties as other Normalizing Flows.

In terms of broader social impacts of this work, generative models of images can be used to generate so-called fake images, and this issue has been discussed at length in other works. We emphasize lower computational budgets, and show comparable performance with far fewer parameters and less training time.


\begin{ack}
Chris Finlay contributed to this paper while a postdoc at McGill University; he is now affiliated with Deep Render. His postdoc was funded in part by a Healthy Brains Healthy Lives Fellowship. Adam Oberman was supported by the Air Force Office of Scientific Research under award number FA9550-18-1-0167 and by IVADO. Christopher Pal is funded in part by CIFAR. We thank CIFAR for their support through the CIFAR AI Chairs program. We also thank Samsung for partially supporting Vikram Voleti for this work. We thank Adam Ibrahim, Etienne Denis, Gauthier Gidel, Ioannis Mitliagkas, and Roger Girgis for their valuable feedback.

\end{ack}


{\small
\bibliographystyle{IEEEtranSN}
\bibliography{ref}

\begin{thebibliography}{83}
\providecommand{\natexlab}[1]{#1}
\providecommand{\url}[1]{#1}
\csname url@samestyle\endcsname
\providecommand{\newblock}{\relax}
\providecommand{\bibinfo}[2]{#2}
\providecommand{\BIBentrySTDinterwordspacing}{\spaceskip=0pt\relax}
\providecommand{\BIBentryALTinterwordstretchfactor}{4}
\providecommand{\BIBentryALTinterwordspacing}{\spaceskip=\fontdimen2\font plus
\BIBentryALTinterwordstretchfactor\fontdimen3\font minus
  \fontdimen4\font\relax}
\providecommand{\BIBforeignlanguage}[2]{{%
\expandafter\ifx\csname l@#1\endcsname\relax
\typeout{** WARNING: IEEEtranSN.bst: No hyphenation pattern has been}%
\typeout{** loaded for the language `#1'. Using the pattern for}%
\typeout{** the default language instead.}%
\else
\language=\csname l@#1\endcsname
\fi
#2}}
\providecommand{\BIBdecl}{\relax}
\BIBdecl

\bibitem[Adelson et~al.(1984)Adelson, Anderson, Bergen, Burt, and
  Ogden]{adelson1984pyramid}
E.~H. Adelson, C.~H. Anderson, J.~R. Bergen, P.~J. Burt, and J.~M. Ogden,
  ``Pyramid methods in image processing,'' \emph{RCA engineer}, vol.~29, no.~6,
  pp. 33--41, 1984.

\bibitem[Arjovsky and Bottou(2017)]{arjovsky2017towards}
M.~Arjovsky and L.~Bottou, ``Towards principled methods for training generative
  adversarial networks,'' \emph{arXiv preprint arXiv:1701.04862}, 2017.

\bibitem[Behrmann et~al.(2019)Behrmann, Grathwohl, Chen, Duvenaud, and
  Jacobsen]{behrmann2019invertible}
J.~Behrmann, W.~Grathwohl, R.~T. Chen, D.~Duvenaud, and J.-H. Jacobsen,
  ``Invertible residual networks,'' in \emph{International Conference on
  Machine Learning}, 2019, pp. 573--582.

\bibitem[Berard et~al.(2020)Berard, Gidel, Almahairi, Vincent, and
  Lacoste-Julien]{berard2019closer}
H.~Berard, G.~Gidel, A.~Almahairi, P.~Vincent, and S.~Lacoste-Julien, ``A
  closer look at the optimization landscapes of generative adversarial
  networks,'' in \emph{International Conference on Machine Learning}, 2020.

\bibitem[Brock et~al.(2019)Brock, Donahue, and Simonyan]{brock2019biggan}
A.~Brock, J.~Donahue, and K.~Simonyan, ``Large scale {GAN} training for high
  fidelity natural image synthesis,'' in \emph{International Conference on
  Learning Representations}, 2019.

\bibitem[Burt and Adelson(1983)]{burt1983laplacian}
P.~Burt and E.~Adelson, ``The laplacian pyramid as a compact image code,''
  \emph{IEEE Transactions on communications}, vol.~31, no.~4, pp. 532--540,
  1983.

\bibitem[Burt(1981)]{burt1981fast}
P.~J. Burt, ``Fast filter transform for image processing,'' \emph{Computer
  graphics and image processing}, vol.~16, no.~1, pp. 20--51, 1981.

\bibitem[Chen et~al.(2020)Chen, Lu, Chenli, Zhu, and Tian]{chen2020vflow}
J.~Chen, C.~Lu, B.~Chenli, J.~Zhu, and T.~Tian, ``Vflow: More expressive
  generative flows with variational data augmentation,'' in \emph{International
  Conference on Machine Learning}, 2020.

\bibitem[Chen et~al.(2018{\natexlab{a}})Chen, Rubanova, Bettencourt, and
  Duvenaud]{chen2018neural}
R.~T.~Q. Chen, Y.~Rubanova, J.~Bettencourt, and D.~Duvenaud, ``Neural ordinary
  differential equations,'' \emph{Advances in Neural Information Processing
  Systems}, 2018.

\bibitem[Chen et~al.(2019)Chen, Behrmann, Duvenaud, and
  Jacobsen]{chen2019residual}
R.~T. Chen, J.~Behrmann, D.~K. Duvenaud, and J.-H. Jacobsen, ``Residual flows
  for invertible generative modeling,'' in \emph{Advances in Neural Information
  Processing Systems}, 2019, pp. 9916--9926.

\bibitem[Chen et~al.(2018{\natexlab{b}})Chen, Mishra, Rohaninejad, and
  Abbeel]{chen2018pixelsnail}
X.~Chen, N.~Mishra, M.~Rohaninejad, and P.~Abbeel, ``Pixelsnail: An improved
  autoregressive generative model,'' in \emph{International Conference on
  Machine Learning}.\hskip 1em plus 0.5em minus 0.4em\relax PMLR, 2018, pp.
  864--872.

\bibitem[Child et~al.(2019)Child, Gray, Radford, and
  Sutskever]{child2019generating}
R.~Child, S.~Gray, A.~Radford, and I.~Sutskever, ``Generating long sequences
  with sparse transformers,'' \emph{arXiv preprint arXiv:1904.10509}, 2019.

\bibitem[Choi et~al.(2018)Choi, Jang, and Alemi]{choi2018waic}
H.~Choi, E.~Jang, and A.~A. Alemi, ``Waic, but why? generative ensembles for
  robust anomaly detection,'' \emph{arXiv preprint arXiv:1810.01392}, 2018.

\bibitem[Deng et~al.(2009)Deng, Dong, Socher, Li, Li, and
  Fei-Fei]{deng2009imagenet}
J.~Deng, W.~Dong, R.~Socher, L.-J. Li, K.~Li, and L.~Fei-Fei, ``Imagenet: A
  large-scale hierarchical image database,'' in \emph{2009 IEEE conference on
  computer vision and pattern recognition}.\hskip 1em plus 0.5em minus
  0.4em\relax IEEE, 2009, pp. 248--255.

\bibitem[Denton et~al.(2015)Denton, Chintala, Fergus, et~al.]{denton2015lapgan}
E.~L. Denton, S.~Chintala, R.~Fergus \emph{et~al.}, ``Deep generative image
  models using a laplacian pyramid of adversarial networks,'' in \emph{Advances
  in neural information processing systems}, 2015, pp. 1486--1494.

\bibitem[Dinh et~al.(2017)Dinh, Sohl-Dickstein, and Bengio]{dinh2016density}
L.~Dinh, J.~Sohl-Dickstein, and S.~Bengio, ``Density estimation using real
  nvp,'' in \emph{International Conference on Learned Representations}, 2017.

\bibitem[Durkan et~al.(2019)Durkan, Bekasov, Murray, and
  Papamakarios]{durkan2019spline}
\BIBentryALTinterwordspacing
C.~Durkan, A.~Bekasov, I.~Murray, and G.~Papamakarios, ``Neural spline flows,''
  in \emph{Advances in Neural Information Processing Systems}, vol.~32, 2019,
  pp. 7511--7522. [Online]. Available:
  \url{https://proceedings.neurips.cc/paper/2019/file/7ac71d433f282034e088473244df8c02-Paper.pdf}
\BIBentrySTDinterwordspacing

\bibitem[Finlay et~al.(2020)Finlay, Jacobsen, Nurbekyan, and
  Oberman]{finlay2020rnode}
C.~Finlay, J.-H. Jacobsen, L.~Nurbekyan, and A.~Oberman, ``How to train your
  neural ode: the world of jacobian and kinetic regularization,''
  \emph{International Conference on Machine Learning}, 2020.

\bibitem[Ghosh et~al.(2020)Ghosh, Behl, Dupont, Torr, and
  Namboodiri]{ghosh2020steer}
A.~Ghosh, H.~S. Behl, E.~Dupont, P.~H. Torr, and V.~Namboodiri, ``Steer: Simple
  temporal regularization for neural odes,'' in \emph{Advances in Neural
  Information Processing Systems}, 2020.

\bibitem[Goodfellow et~al.(2016)Goodfellow, Bengio, Courville, and
  Bengio]{goodfellow2016deep}
I.~Goodfellow, Y.~Bengio, A.~Courville, and Y.~Bengio, \emph{Deep
  learning}.\hskip 1em plus 0.5em minus 0.4em\relax MIT Press, 2016, vol.~1.

\bibitem[Grathwohl et~al.(2019)Grathwohl, Chen, Bettencourt, Sutskever, and
  Duvenaud]{grathwohl2019ffjord}
W.~Grathwohl, R.~T.~Q. Chen, J.~Bettencourt, I.~Sutskever, and D.~Duvenaud,
  ``Ffjord: Free-form continuous dynamics for scalable reversible generative
  models,'' \emph{International Conference on Learning Representations}, 2019.

\bibitem[Grcić et~al.(2021)Grcić, Grubišić, and
  Šegvić]{grcic2021denseflow}
\BIBentryALTinterwordspacing
M.~Grcić, I.~Grubišić, and S.~Šegvić, ``Densely connected normalizing
  flows,'' \emph{arXiv preprint}, 2021. [Online]. Available: \url{2010.02502}
\BIBentrySTDinterwordspacing

\bibitem[Hendrycks and Gimpel(2017)]{hendrycks2017baseline}
D.~Hendrycks and K.~Gimpel, ``A baseline for detecting misclassified and
  out-of-distribution examples in neural networks,'' in \emph{International
  Conference on Learning Representations}, 2017.

\bibitem[Hendrycks et~al.(2019)Hendrycks, Mazeika, and
  Dietterich]{hendrycks2019deep}
D.~Hendrycks, M.~Mazeika, and T.~Dietterich, ``Deep anomaly detection with
  outlier exposure,'' in \emph{International Conference on Learning
  Representations}, 2019.

\bibitem[Ho et~al.(2019{\natexlab{a}})Ho, Chen, Srinivas, Duan, and
  Abbeel]{ho2019flow++}
J.~Ho, X.~Chen, A.~Srinivas, Y.~Duan, and P.~Abbeel, ``Flow++: Improving
  flow-based generative models with variational dequantization and architecture
  design,'' in \emph{International Conference on Machine Learning}, 2019.

\bibitem[Ho et~al.(2019{\natexlab{b}})Ho, Kalchbrenner, Weissenborn, and
  Salimans]{ho2019axial}
J.~Ho, N.~Kalchbrenner, D.~Weissenborn, and T.~Salimans, ``Axial attention in
  multidimensional transformers,'' \emph{arXiv preprint arXiv:1912.12180},
  2019.

\bibitem[Hoogeboom et~al.(2019{\natexlab{a}})Hoogeboom, Berg, and
  Welling]{hoogeboom2019emerging}
E.~Hoogeboom, R.~v.~d. Berg, and M.~Welling, ``Emerging convolutions for
  generative normalizing flows,'' in \emph{International Conference on Machine
  Learning}, 2019.

\bibitem[Hoogeboom et~al.(2019{\natexlab{b}})Hoogeboom, Peters, van~den Berg,
  and Welling]{hoogeboom2019idf}
\BIBentryALTinterwordspacing
E.~Hoogeboom, J.~Peters, R.~van~den Berg, and M.~Welling, ``Integer discrete
  flows and lossless compression,'' in \emph{Advances in Neural Information
  Processing Systems}, vol.~32, 2019, pp. 12\,134--12\,144. [Online].
  Available:
  \url{https://proceedings.neurips.cc/paper/2019/file/9e9a30b74c49d07d8150c8c83b1ccf07-Paper.pdf}
\BIBentrySTDinterwordspacing

\bibitem[Hoogeboom et~al.(2020)Hoogeboom, , Satorras, Tomczak, and
  Welling]{hoogeboom2020flow}
E.~Hoogeboom, , V.~G. Satorras, J.~Tomczak, and M.~Welling, ``The convolution
  exponential and generalized sylvester flows,'' in \emph{Advances in Neural
  Information Processing Systems}, 2020.

\bibitem[H{\o}st-Madsen et~al.(2019)H{\o}st-Madsen, Sabeti, and
  Walton]{host2019data}
A.~H{\o}st-Madsen, E.~Sabeti, and C.~Walton, ``Data discovery and anomaly
  detection using atypicality: Theory,'' \emph{IEEE Transactions on Information
  Theory}, vol.~65, no.~9, pp. 5302--5322, 2019.

\bibitem[Huang et~al.(2020)Huang, Dinh, and Courville]{huang2020augmented}
C.-W. Huang, L.~Dinh, and A.~Courville, ``Augmented normalizing flows: Bridging
  the gap between generative flows and latent variable models,'' \emph{arXiv
  preprint arXiv:2002.07101}, 2020.

\bibitem[Huang and Yeh(2021)]{huang2021acc}
H.-H. Huang and M.-Y. Yeh, ``Accelerating continuous normalizing flow with
  trajectory polynomial regularization,'' \emph{AAAI Conference on Artificial
  Intelligence}, 2021.

\bibitem[Jimenez~Rezende and Mohamed(2015)]{jimenez2015variational}
D.~Jimenez~Rezende and S.~Mohamed, ``Variational inference with normalizing
  flows,'' in \emph{International Conference on Machine Learning}, 2015, pp.
  1530–--1538.

\bibitem[Jun et~al.(2020)Jun, Child, Chen, Schulman, Ramesh, Radford, and
  Sutskever]{jun2020distaug}
H.~Jun, R.~Child, M.~Chen, J.~Schulman, A.~Ramesh, A.~Radford, and
  I.~Sutskever, ``Distribution augmentation for generative modeling,'' in
  \emph{International Conference on Machine Learning}, 2020, pp.
  10\,563--10\,576.

\bibitem[Karami et~al.(2019)Karami, Schuurmans, Sohl-Dickstein, Dinh, and
  Duckworth]{karami2019invertible}
\BIBentryALTinterwordspacing
M.~Karami, D.~Schuurmans, J.~Sohl-Dickstein, L.~Dinh, and D.~Duckworth,
  ``Invertible convolutional flow,'' in \emph{Advances in Neural Information
  Processing Systems}, vol.~32, 2019, pp. 5635--5645. [Online]. Available:
  \url{https://proceedings.neurips.cc/paper/2019/file/b1f62fa99de9f27a048344d55c5ef7a6-Paper.pdf}
\BIBentrySTDinterwordspacing

\bibitem[Karnewar and Wang(2020)]{karnewar2020msg}
A.~Karnewar and O.~Wang, ``Msg-gan: Multi-scale gradients for generative
  adversarial networks,'' in \emph{Proceedings of the IEEE/CVF Conference on
  Computer Vision and Pattern Recognition}, 2020, pp. 7799--7808.

\bibitem[Karras et~al.(2018)Karras, Aila, Laine, and
  Lehtinen]{karras2017progressive}
T.~Karras, T.~Aila, S.~Laine, and J.~Lehtinen, ``Progressive growing of gans
  for improved quality, stability, and variation,'' in \emph{International
  Conference on Learned Representations}, 2018.

\bibitem[Karras et~al.(2020)Karras, Laine, Aittala, Hellsten, Lehtinen, and
  Aila]{karras2020analyzing}
T.~Karras, S.~Laine, M.~Aittala, J.~Hellsten, J.~Lehtinen, and T.~Aila,
  ``Analyzing and improving the image quality of stylegan,'' in
  \emph{Proceedings of the IEEE/CVF Conference on Computer Vision and Pattern
  Recognition}, 2020, pp. 8110--8119.

\bibitem[Kelly et~al.(2020)Kelly, Bettencourt, Johnson, and
  Duvenaud]{kelly2020learning}
J.~Kelly, J.~Bettencourt, M.~J. Johnson, and D.~Duvenaud, ``Learning
  differential equations that are easy to solve,'' in \emph{Advances in Neural
  Information Processing Systems}, 2020.

\bibitem[Kingma and Welling(2013)]{kingma2013auto}
D.~P. Kingma and M.~Welling, ``Auto-encoding variational bayes,'' \emph{arXiv
  preprint arXiv:1312.6114}, 2013.

\bibitem[Kingma et~al.(2016)Kingma, Salimans, Jozefowicz, Chen, Sutskever, and
  Welling]{kingma2016improved}
\BIBentryALTinterwordspacing
D.~P. Kingma, T.~Salimans, R.~Jozefowicz, X.~Chen, I.~Sutskever, and
  M.~Welling, ``Improving variational inference with inverse autoregressive
  flow,'' 2016, cite arxiv:1606.04934. [Online]. Available:
  \url{http://arxiv.org/abs/1606.04934}
\BIBentrySTDinterwordspacing

\bibitem[Kingma and Dhariwal(2018)]{kingma2018glow}
D.~P. Kingma and P.~Dhariwal, ``Glow: Generative flow with invertible 1x1
  convolutions,'' in \emph{Advances in neural information processing systems},
  2018, pp. 10\,215--10\,224.

\bibitem[Kirichenko et~al.(2020)Kirichenko, Izmailov, and
  Wilson]{kirichenko2020normalizing}
P.~Kirichenko, P.~Izmailov, and A.~G. Wilson, ``Why normalizing flows fail to
  detect out-of-distribution data,'' in \emph{Advances in neural information
  processing systems}, vol.~33, 2020.

\bibitem[Kobyzev et~al.(2020)Kobyzev, Prince, and
  Brubaker]{kobyzev2020normalizing}
I.~Kobyzev, S.~Prince, and M.~Brubaker, ``Normalizing flows: An introduction
  and review of current methods,'' \emph{IEEE Transactions on Pattern Analysis
  and Machine Intelligence}, 2020.

\bibitem[Krizhevsky et~al.(2009)Krizhevsky, Hinton,
  et~al.]{krizhevsky2009learning}
\BIBentryALTinterwordspacing
A.~Krizhevsky, G.~Hinton \emph{et~al.}, ``Learning multiple layers of features
  from tiny images,'' \emph{Technical Report, University of Toronto}, 2009.
  [Online]. Available: \url{https://www.cs.toronto.edu/~kriz/cifar.html}
\BIBentrySTDinterwordspacing

\bibitem[Lee et~al.(2018)Lee, Lee, Lee, and Shin]{lee2018simple}
K.~Lee, K.~Lee, H.~Lee, and J.~Shin, ``A simple unified framework for detecting
  out-of-distribution samples and adversarial attacks,'' in \emph{Advances in
  Neural Information Processing Systems}, 2018, pp. 7167--7177.

\bibitem[Lee et~al.(2020)Lee, Kim, and Yoon]{lee2020nanoflow}
S.-g. Lee, S.~Kim, and S.~Yoon, ``Nanoflow: Scalable normalizing flows with
  sublinear parameter complexity,'' in \emph{Advances in Neural Information
  Processing Systems}, 2020.

\bibitem[Liang et~al.(2018)Liang, Li, and Srikant]{liang2018enhancing}
S.~Liang, Y.~Li, and R.~Srikant, ``Enhancing the reliability of
  out-of-distribution image detection in neural networks,'' in
  \emph{International Conference on Learning Representations}, 2018.

\bibitem[Lin et~al.(2018)Lin, Khetan, Fanti, and Oh]{lin2018pacgan}
Z.~Lin, A.~Khetan, G.~Fanti, and S.~Oh, ``Pacgan: The power of two samples in
  generative adversarial networks,'' in \emph{Advances in neural information
  processing systems}, 2018, pp. 1498--1507.

\bibitem[Lindeberg(1990)]{lindeberg1990scale}
T.~Lindeberg, ``Scale-space for discrete signals,'' \emph{IEEE transactions on
  pattern analysis and machine intelligence}, vol.~12, no.~3, pp. 234--254,
  1990.

\bibitem[Long et~al.(2015)Long, Shelhamer, and Darrell]{long2015fully}
J.~Long, E.~Shelhamer, and T.~Darrell, ``Fully convolutional networks for
  semantic segmentation,'' in \emph{Proceedings of the IEEE conference on
  computer vision and pattern recognition}, 2015, pp. 3431--3440.

\bibitem[Lu and Huang(2020)]{lu2020woodbury}
Y.~Lu and B.~Huang, ``Woodbury transformations for deep generative flows,'' in
  \emph{Advances in Neural Information Processing Systems}, 2020.

\bibitem[Ma et~al.(2019)Ma, Kong, Zhang, and Hovy]{ma2019macow}
X.~Ma, X.~Kong, S.~Zhang, and E.~Hovy, ``Macow: Masked convolutional generative
  flow,'' in \emph{Advances in Neural Information Processing Systems}, 2019,
  pp. 5893--5902.

\bibitem[Mallat(1989)]{mallat1989theory}
S.~G. Mallat, ``A theory for multiresolution signal decomposition: the wavelet
  representation,'' \emph{IEEE transactions on pattern analysis and machine
  intelligence}, vol.~11, no.~7, pp. 674--693, 1989.

\bibitem[Mallat and Peyr\'e(2009)]{mallat2009wavelet}
S.~G. Mallat and G.~Peyr\'e, \emph{A wavelet tour of signal processing: the
  sparse way}.\hskip 1em plus 0.5em minus 0.4em\relax Elsevier, 2009.

\bibitem[Marr(2010)]{marr2010vision}
D.~Marr, \emph{Vision: A computational investigation into the human
  representation and processing of visual information}.\hskip 1em plus 0.5em
  minus 0.4em\relax MIT press, 2010.

\bibitem[Menick and Kalchbrenner(2019)]{menick2019generating}
J.~Menick and N.~Kalchbrenner, ``Generating high fidelity images with subscale
  pixel networks and multidimensional upscaling,'' in \emph{International
  Conference on Learning Representations}, 2019.

\bibitem[Nalisnick et~al.(2019{\natexlab{a}})Nalisnick, Matsukawa, Teh, Gorur,
  and Lakshminarayanan]{nalisnick2018deep}
E.~Nalisnick, A.~Matsukawa, Y.~W. Teh, D.~Gorur, and B.~Lakshminarayanan, ``Do
  deep generative models know what they don't know?'' in \emph{International
  Conference on Learning Representations}, 2019.

\bibitem[Nalisnick et~al.(2019{\natexlab{b}})Nalisnick, Matsukawa, Teh, and
  Lakshminarayanan]{nalisnick2019detecting}
E.~Nalisnick, A.~Matsukawa, Y.~W. Teh, and B.~Lakshminarayanan, ``Detecting
  out-of-distribution inputs to deep generative models using a test for
  typicality,'' \emph{arXiv preprint arXiv:1906.02994}, vol.~5, 2019.

\bibitem[Netzer et~al.(2011)Netzer, Wang, Coates, Bissacco, Wu, and
  Ng]{Netzer2011svhn}
Y.~Netzer, T.~Wang, A.~Coates, A.~Bissacco, B.~Wu, and A.~Y. Ng, ``Reading
  digits in natural images with unsupervised feature learning,'' \emph{NIPS
  Workshop on Deep Learning and Unsupervised Feature Learning}, 2011.

\bibitem[Nielsen and Winther(2020)]{nielsen2020closing}
D.~Nielsen and O.~Winther, ``Closing the dequantization gap: Pixelcnn as a
  single-layer flow,'' in \emph{Advances in Neural Information Processing
  Systems}, 2020.

\bibitem[Onken et~al.(2021)Onken, Fung, Li, and Ruthotto]{onken2021otflow}
D.~Onken, S.~W. Fung, X.~Li, and L.~Ruthotto, ``Ot-flow: Fast and accurate
  continuous normalizing flows via optimal transport,'' \emph{AAAI Conference
  on Artificial Intelligence}, 2021.

\bibitem[Oord et~al.(2016)Oord, Kalchbrenner, and Kavukcuoglu]{oord2016pixel}
A.~v.~d. Oord, N.~Kalchbrenner, and K.~Kavukcuoglu, ``Pixel recurrent neural
  networks,'' \emph{International Conference on Machine Learning}, 2016.

\bibitem[Papamakarios et~al.(2019)Papamakarios, Nalisnick, Rezende, Mohamed,
  and Lakshminarayanan]{papamakarios2019normalizing}
G.~Papamakarios, E.~Nalisnick, D.~J. Rezende, S.~Mohamed, and
  B.~Lakshminarayanan, ``Normalizing flows for probabilistic modeling and
  inference,'' \emph{arXiv preprint arXiv:1912.02762}, 2019.

\bibitem[Parmar et~al.(2018)Parmar, Vaswani, Uszkoreit, Kaiser, Shazeer, Ku,
  and Tran]{parmar2018image}
N.~Parmar, A.~Vaswani, J.~Uszkoreit, {\L}.~Kaiser, N.~Shazeer, A.~Ku, and
  D.~Tran, ``Image transformer,'' in \emph{International Conference on Machine
  Learning}, 2018.

\bibitem[Pascanu et~al.(2013)Pascanu, Mikolov, and
  Bengio]{pascanu2013difficulty}
R.~Pascanu, T.~Mikolov, and Y.~Bengio, ``On the difficulty of training
  recurrent neural networks,'' in \emph{International conference on machine
  learning}, 2013, pp. 1310--1318.

\bibitem[Radford et~al.(2015)Radford, Metz, and
  Chintala]{radford2015unsupervised}
A.~Radford, L.~Metz, and S.~Chintala, ``Unsupervised representation learning
  with deep convolutional generative adversarial networks,'' \emph{arXiv
  preprint arXiv:1511.06434}, 2015.

\bibitem[Razavi et~al.(2019)Razavi, van~den Oord, and
  Vinyals]{razavi2019generating}
A.~Razavi, A.~van~den Oord, and O.~Vinyals, ``Generating diverse high-fidelity
  images with vq-vae-2,'' in \emph{Advances in Neural Information Processing
  Systems}, 2019, pp. 14\,866--14\,876.

\bibitem[Reed et~al.(2017)Reed, Oord, Kalchbrenner, Colmenarejo, Wang, Belov,
  and De~Freitas]{reed2017parallel}
S.~Reed, A.~v.~d. Oord, N.~Kalchbrenner, S.~G. Colmenarejo, Z.~Wang, D.~Belov,
  and N.~De~Freitas, ``Parallel multiscale autoregressive density estimation,''
  in \emph{International Conference on Machine Learning}, 2017.

\bibitem[Sabeti and H{\o}st-Madsen(2019)]{sabeti2019data}
E.~Sabeti and A.~H{\o}st-Madsen, ``Data discovery and anomaly detection using
  atypicality for real-valued data,'' \emph{Entropy}, vol.~21, no.~3, p. 219,
  2019.

\bibitem[Serr{\`a} et~al.(2020)Serr{\`a}, {\'A}lvarez, G{\'o}mez, Slizovskaia,
  N{\'u}{\~n}ez, and Luque]{serra2019input}
J.~Serr{\`a}, D.~{\'A}lvarez, V.~G{\'o}mez, O.~Slizovskaia, J.~F.
  N{\'u}{\~n}ez, and J.~Luque, ``Input complexity and out-of-distribution
  detection with likelihood-based generative models,'' in \emph{International
  Conference on Learning Representations}, 2020.

\bibitem[Shaham et~al.(2019)Shaham, Dekel, and Michaeli]{shaham2019singan}
T.~R. Shaham, T.~Dekel, and T.~Michaeli, ``Singan: Learning a generative model
  from a single natural image,'' in \emph{Proceedings of the IEEE International
  Conference on Computer Vision}, 2019, pp. 4570--4580.

\bibitem[Sneyers and Wuille(2016)]{sneyers2016flif}
J.~Sneyers and P.~Wuille, ``Flif: Free lossless image format based on maniac
  compression,'' in \emph{2016 IEEE International Conference on Image
  Processing (ICIP)}.\hskip 1em plus 0.5em minus 0.4em\relax IEEE, 2016, pp.
  66--70.

\bibitem[Song et~al.(2019)Song, Meng, and Ermon]{song2019mintnet}
Y.~Song, C.~Meng, and S.~Ermon, ``Mintnet: Building invertible neural networks
  with masked convolutions,'' in \emph{Advances in Neural Information
  Processing Systems}, 2019, pp. 11\,004--11\,014.

\bibitem[Tabak and Turner(2013)]{tabak2013family}
E.~G. Tabak and C.~V. Turner, ``A family of nonparametric density estimation
  algorithms,'' \emph{Communications on Pure and Applied Mathematics}, vol.~66,
  no.~2, pp. 145--164, 2013.

\bibitem[Theis et~al.(2016)Theis, Oord, and Bethge]{theis2016note}
L.~Theis, A.~v.~d. Oord, and M.~Bethge, ``A note on the evaluation of
  generative models,'' in \emph{International Conference on Learning
  Representations}, 2016.

\bibitem[Vahdat and Kautz(2020)]{vahdat2020nvae}
A.~Vahdat and J.~Kautz, ``Nvae: A deep hierarchical variational autoencoder,''
  in \emph{Advances in Neural Information Processing Systems}, 2020.

\bibitem[Van~den Oord et~al.(2016)Van~den Oord, Kalchbrenner, Espeholt,
  Vinyals, Graves, et~al.]{van2016conditional}
A.~Van~den Oord, N.~Kalchbrenner, L.~Espeholt, O.~Vinyals, A.~Graves
  \emph{et~al.}, ``Conditional image generation with pixelcnn decoders,'' in
  \emph{Advances in neural information processing systems}, 2016, pp.
  4790--4798.

\bibitem[Varga(1966)]{varga1966discrete}
R.~S. Varga, ``On a discrete maximum principle,'' \emph{SIAM Journal on
  Numerical Analysis}, vol.~3, no.~2, pp. 355--359, 1966.

\bibitem[Witkin(1987)]{witkin1987scale}
A.~P. Witkin, ``Scale-space filtering,'' in \emph{Readings in Computer
  Vision}.\hskip 1em plus 0.5em minus 0.4em\relax Elsevier, 1987, pp. 329--332.

\bibitem[Xiao and Liu(2020)]{xiao2020generative}
C.~Xiao and L.~Liu, ``Generative flows with matrix exponential,'' in
  \emph{International Conference on Machine Learning}, 2020.

\bibitem[Yan et~al.(2020)Yan, Du, Tan, and Feng]{yan2020robustness}
H.~Yan, J.~Du, V.~Y.~F. Tan, and J.~Feng, ``On robustness of neural ordinary
  differential equations,'' \emph{International Conference on Learning
  Representations}, 2020.

\bibitem[Yu et~al.(2020)Yu, Derpanis, and Brubaker]{yu2020waveletflow}
J.~Yu, K.~Derpanis, and M.~Brubaker, ``Wavelet flow: Fast training of high
  resolution normalizing flows,'' in \emph{Advances in Neural Information
  Processing Systems}, 2020.

\end{thebibliography}
}

\clearpage
\appendix

\section{Full Table 1}
\begin{table*}[!h]
    \small
    \centering
    \caption{\normalsize{Unconditional image generation metrics (lower is better in all cases): number of parameters in the model, bits-per-dimension, time (in hours). Most previous models use multiple GPUs for training, all our models were trained on only \textit{one} NVIDIA V100 GPU.
    $^\ddagger$As reported in~\cite{ghosh2020steer}.
    $^*$FFJORD RNODE~\cite{finlay2020rnode} used 4 GPUs to train on ImageNet64. `x': Fails to train.}}
    {\setlength{\tabcolsep}{.2em}
    \begin{tabular}{|l|crc|rrr|crc|}
        \hline
        & \multicolumn{3}{c|}{\textbf{\textsc{CIFAR10}}} & \multicolumn{3}{c|}{\textbf{\textsc{ImageNet32}}} & \multicolumn{3}{c|}{\textbf{\textsc{ImageNet64}}} \\
        & \textsc{BPD} & \textsc{Param} & \textsc{Time}
        & \textsc{BPD} & \textsc{Param} & \textsc{Time}
        & \textsc{BPD} & \textsc{Param} & \textsc{Time} \\
        \hline\hline
        \multicolumn{10}{|l|}{\textbf{Non Flow-based Prior Work}} \\
        \hline
        PixelRNN~\citep{oord2016pixel} 
            & 3.00 & & 
            & 3.86 & & 
            & 3.63 & & \\[-2pt]
        Gated PixelCNN
        ~\citep{van2016conditional}
            & 3.03 &  & 
            & 3.83 &  & 60
            & 3.57 &  & 60 \\[-2pt]
        Parallel Multiscale
        ~\citep{reed2017parallel}
            & & & 
            & 3.95 & & 
            & 3.70 & &  \\[-2pt]
        Image Transformer
        ~\cite{parmar2018image}
            & 2.90 &  & 
            & 3.77 &  &  
            &  &  &   \\[-2pt]
        PixelSNAIL
        ~\citep{chen2018pixelsnail}
            & 2.85 &  & 
            & 3.80 &  &  
            &  &  &   \\[-2pt]
        SPN
        ~\citep{menick2019generating}
            &  &  & 
            & 3.85 & \footnotesize{150.0M} &  
            & 3.53 & \footnotesize{150.0M} &   \\[-2pt]
        Sparse Transformer
        ~\citep{child2019generating}
            & 2.80 & \footnotesize{59.0M} & 
            &  &  &  
            & 3.44 & \footnotesize{152.0M} & \footnotesize{7days}  \\[-2pt]
        Axial Transformer
        ~\citep{ho2019axial}
            &  &  & 
            & 3.76 &  &  
            & 3.44 &  &   \\[-2pt]
        PixelFlow++
        ~\citep{nielsen2020closing}
            & 2.92 &  &
            &  &  & 
            &  &  &  \\[-2pt]
        NVAE
        ~\citep{vahdat2020nvae}
            & 2.91 &  & 55
            & 3.92 &  & 70 
            &  &  &   \\[-2pt]
        Dist-Aug
        Sparse Transformer
        ~\citep{jun2020distaug}
            & 2.56 & \footnotesize{152.0M} &
            &  &  & 
            & 3.42 & \footnotesize{152.0M} &  \\
        \hline
        \hline
        \multicolumn{10}{|l|}{\textbf{Flow-based Prior Work}} \\
        \hline
        IAF
        ~\citep{kingma2016improved}
            &  &  & 
            & 3.11 & &
            &  &  &  \\
        RealNVP
        ~\citep{dinh2016density}
            & 3.49 & & 
            & 4.28 & \footnotesize{46.0M} & 
            & 3.98 & \footnotesize{96.0M} & \\[-2pt]
        Glow
        ~\citep{kingma2018glow}
            & 3.35 & \footnotesize{44.0M} &
            & 4.09 & \footnotesize{66.1M} & 
            & 3.81 & \footnotesize{111.1M} & \\[-2pt]
        i-ResNets
        ~\citep{behrmann2019invertible}
            &  &  & 
            &  &  &  
            &  &  &  \\[-2pt]
        Emerging
        ~\citep{hoogeboom2019emerging}
            & 3.34 & \footnotesize{44.7M} &
            & 4.09 & \footnotesize{67.1M} &
            & 3.81 & \footnotesize{67.1M} & \\[-2pt]
        IDF
        ~\citep{hoogeboom2019idf}
            & 3.34 & &
            & 4.18 & &
            & 3.90 & & \\[-2pt]
        S-CONF
        ~\citep{karami2019invertible}
            & 3.34 & &
            & & &
            & & & \\[-2pt]
        MintNet
        ~\citep{song2019mintnet}
            & 3.32 & \footnotesize{17.9M} & \footnotesize{$\geq$5days}
            & 4.06 & \footnotesize{17.4M} &
            &  &  &  \\[-2pt]
        Residual Flow
        ~\citep{chen2019residual}
            & 3.28 & &
            & 4.01 & &
            & 3.76 & & \\[-2pt]
        MaCow
        ~\citep{ma2019macow}
            & 3.16 & \footnotesize{43.5M} &
            &  &  & 
            & 3.69 & \footnotesize{122.5M} & \\[-2pt]
        Neural Spline Flows
        ~\citep{durkan2019spline}
            & 3.38 & \footnotesize{11.8M} &
            & & &
            & 3.82 & \footnotesize{15.6M} & \\[-2pt]
        Flow++
        ~\citep{ho2019flow++}
            & 3.08 & \footnotesize{31.4M} &
            & 3.86  & \footnotesize{169.0M} &
            & 3.69 & \footnotesize{73.5M} & \\[-2pt]
        ANF
        ~\citep{huang2020augmented}
            & 3.05 & &
            & 3.92 & &
            & 3.66 & & \\[-2pt]
        MEF
        ~\citep{xiao2020generative}
            & 3.32 & \footnotesize{37.7M} &
            & 4.05 & \footnotesize{37.7M} &
            & 3.73 & \footnotesize{46.6M} & \\[-2pt]
        VFlow
        ~\citep{chen2020vflow}
            & 2.98 & & 
            & 3.83 & & 
            &  &  &   \\[-2pt]
        Woodbury NF
        ~\citep{lu2020woodbury}
            & 3.47 & & 
            & 4.20 & & 
            & 3.87 & &  \\
        NanoFlow
        ~\citep{lee2020nanoflow}
            &  3.25 & & 
            &  &  &  
            &  &  &   \\
        ConvExp
        ~\citep{hoogeboom2020flow}
            &  3.218 & &
            &  &  & 
            &  &  &  \\
        Wavelet Flow
        ~\citep{yu2020waveletflow}
            &  &  & 
            & 4.08 & \footnotesize{64.0M} & 
            & 3.78 & \footnotesize{96.0M} & 822  \\
        TayNODE
        ~\citep{kelly2020learning}
            & 1.039 & & 
            & & & 
            & & &  \\
        \hline
        \hline
        \multicolumn{10}{|l|}{\textbf{1-resolution Continuous Normalizing Flow}}\\
        \hline
        FFJORD
        ~\citep{grathwohl2019ffjord}
            & 3.40 & 0.9M & \footnotesize{$\geq$5days}
            & $^{\ddagger}3.96$ & $^{\ddagger}$\footnotesize{2.0M} & $^{\ddagger}$\footnotesize{$>$5days}
            & x &  & x \\[-0pt]
        RNODE
        ~\citep{finlay2020rnode}
            & 3.38 & 1.4M & 31.84
            & $^{\ddagger}2.36$ & \footnotesize{2.0M} & $^{\ddagger}30.1$
            & $^*3.83$ & \footnotesize{2.0M} & $^*256.4$ \\[-3pt]
            &  &  & 
            & $^{\mathsection}3.49$ & $^\mathsection$\footnotesize{1.6M} & $^{\mathsection}40.39$
            & & &  \\
        FFJORD + STEER
        ~\citep{ghosh2020steer}
            & 3.40 & \footnotesize{1.4M} & 86.34
            & 3.84 & \footnotesize{2.0M} & \footnotesize{$>$5days}
            & & & \\
        RNODE + STEER
        ~\citep{ghosh2020steer}
            & 3.397 & \footnotesize{1.4M} & 22.24
            & 2.35 & \footnotesize{2.0M} & 24.90
            & & & \\[-3pt]
            & & & 
            & $^\mathsection3.49$ & $^\mathsection$\footnotesize{1.6M} & $^\mathsection30.07$
            & & & \\
        \hline
        \hline
        \multicolumn{10}{|l|}{\textbf{\textsc{(Ours)} Multi-Resolution Continuous Normalizing Flow (MRCNF)}}\\
        \hline
        2-resolution MRCNF
            & 3.65 & \footnotesize{1.3M} & 19.79
            & 3.77 & \footnotesize{1.3M} & 18.18
            & 3.44& \footnotesize{2.0M} & 42.30 \\
        2-resolution MRCNF
            & 3.54 & \footnotesize{3.3M} & 36.47
            & 3.78 & \footnotesize{6.7M} & 17.98
            & x & \footnotesize{6.7M} & x \\
        3-resolution MRCNF
            & 3.79 & \footnotesize{1.5M} & 17.44
            & 3.97 & \footnotesize{1.5M} & 13.78
            & 3.55 & \footnotesize{2.0M} & 35.39 \\
        3-resolution MRCNF
            & 3.60 & \footnotesize{5.1M} & 38.27
            & 3.93 & \footnotesize{10.2M} & 41.20
            & x & \footnotesize{7.6M} & x\\
        \hline
    \end{tabular}
    \label{tab:msflow}
    \vspace{-4em}
    }
\end{table*}

\section{Qualitative samples}

\begin{figure}[!htb]
    \centering
    \begin{subfigure}[b]{0.45\textwidth}
     \centering
     \includegraphics[width=\textwidth]{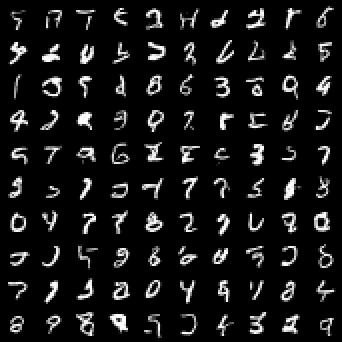}
     \caption{Generated samples at $16\x16$}
     \label{fig:mnist16}
    \end{subfigure}
    \hfill
    \begin{subfigure}[b]{0.45\textwidth}
     \centering
     \includegraphics[width=\textwidth]{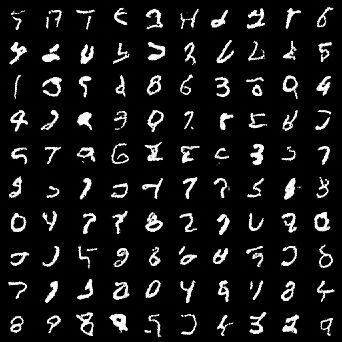}
     \caption{Corresponding generated samples at $32\x32$}
     \label{fig:mnist32}
    \end{subfigure}
    \caption{\small{Generated samples from MNIST.}}
    \label{fig:images-mnist}
\end{figure}

\begin{figure}[!htb]
    \centering
    \begin{subfigure}[b]{0.3\textwidth}
     \centering
     \includegraphics[width=\textwidth]{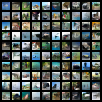}
     \caption{Generated samples at $8\x8$}
    \end{subfigure}
    \hfill
    \begin{subfigure}[b]{0.3\textwidth}
     \centering
     \includegraphics[width=\textwidth]{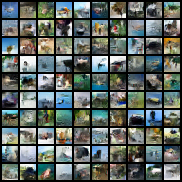}
     \caption{Generated samples at $16\x16$}
    \end{subfigure}
    \hfill
    \begin{subfigure}[b]{0.3\textwidth}
     \centering
     \includegraphics[width=\textwidth]{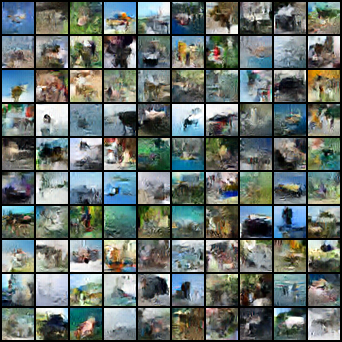}
     \caption{Generated samples at $32\x32$}
    \end{subfigure}
    \caption{\small{Generated samples from CIFAR10.}}
    \label{fig:cifar10}
\end{figure}


\section{Simple example of density estimation}

For example, if we use Euler method as our ODE solver, for density estimation \autoref{eq:neural_ode} reduces to:
\begin{align}
\rvv(t_1) = \rvv(t_0) + (t_1 - t_0)f_s(\rvv(t_0), t_0  \mid \rvc)
\end{align}
where $f_s$ is a neural network, $t_0$ represents the "time" at which the state is image $\rvx$, and $t_1$ is when the state is noise $\rvz$.
We start at scale S with an image sample $\rvx_S$, and assume $t_0$ and $t_1$ are $0$ and $1$ respectively:
\begin{align}
\begin{cases}
&\rvz_S = \rvx_S + f_S(\rvx_S,\ t_0 \mid \rvx_{S-1})\\
&\rvz_{S-1} = \rvx_{S-1} + f_{S-1}(\rvx_{S-1},\ t_0  \mid \rvx_{S-2})\\
&\vdots\\
&\rvz_1 = \rvx_1 + f_1(\rvx_1,\ t_0 \mid \rvx_0)\\
&\rvz_0 = \rvx_0 + f_0(\rvx_0,\ t_0)
\end{cases}
\label{eq:density_ex}
\end{align}

\section{Simple example of generation}

For example, if we use Euler method as our ODE solver, for generation \autoref{eq:neural_ode} reduces to:
\begin{align}
\rvv(t_0) = \rvv(t_1) + (t_0 - t_1)f_s(\rvv(t_1), t_1 \mid \rvc)
\end{align}
i.e. the state is integrated backwards from $t_1$ (i.e. $\rvz_s$) to $t_0$ (i.e. $\rvx_s$).
We start at scale 0 with a noise sample $\rvz_0$, and assume $t_0$ and $t_1$ are $0$ and $1$ respectively:
\begin{align}
\begin{cases}
&\rvx_0 = \rvz_0 - f_0(\rvz_0,\ t_1)\\
&\rvx_1 = \rvz_1 - f_1(\rvz_1,\ t_1 \mid \rvx_0)\\
&\vdots\\
&\rvx_{S-1} = \rvz_{S-1} - f_{S-1}(\rvz_{S-1},\ t_1 \mid \rvx_{S-2})\\
&\rvx_S = \rvz_S - f_S(\rvz_{S},\ t_1 \mid \rvx_{S-1})
\end{cases}
\label{eq:generation_ex}
\end{align}

\section{Models}

We used the same neural network architecture as in RNODE~\cite{finlay2020rnode}. The CNF at each resolution consists of a stack of $bl$ blocks of a 4-layer deep convolutional network comprised of 3x3 kernels and softplus activation functions, with 64 hidden dimensions, and time t concatenated to the spatial input. In addition, except at the coarsest resolution, the immediate coarser image is also concatenated with the state. The integration time of each piece is [0, 1]. The number of blocks $bl$ and the corresponding total number of parameters are given in \autoref{tab:cifar_params}.

\begin{table}[htb]
\setlength{\tabcolsep}{2pt}
\centering
\caption{Number of parameters for different models with different total number of resolutions (res), and the number of channels (ch) and number of blocks (bl) per resolution.}
\begin{tabular}{ccc|c}
\multicolumn{4}{c}{\gls{msflow-image}}\\
\hline
resolutions & ch & bl & Param\\
\hline
\multirow{3}{*}{1}
& 64 & 2 & \footnotesize{0.16M}\\[-0pt]
& 64 & 4 & \footnotesize{0.32M}\\[-0pt]
& 64 & 14 & \footnotesize{1.10M}\\
\hline
\multirow{3}{*}{2}
& 64 & 8 & \footnotesize{1.33M}\\[-0pt]
& 64 & 20 & \footnotesize{3.34M}\\[-0pt]
& 64 & 40 & \footnotesize{6.68M}\\[-0pt]
\hline
\multirow{3}{*}{3}
& 64 & 6 & \footnotesize{1.53M}\\[-0pt]
& 64 & 8 & \footnotesize{2.04M}\\[-0pt]
& 64 & 20 & \footnotesize{5.10M}\\[-0pt]
\end{tabular}
\label{tab:cifar_params}
\vspace{-0.5em}
\end{table}

\section{Gradient norm}

In order to avoid exploding gradients, We clipped the norm of the gradients~\cite{pascanu2013difficulty} by a maximum value of 100.0. In case of using adversarial loss, we first clip the gradients provided by the adversarial loss by 50.0, sum up the gradients provided by the log-likelihood loss, and then clip the summed gradients by 100.0.

\section{8-bit to uniform}
The change-of-variables formula gives the change in probability due to the transformation of $\rvu$ to $\rvv$:
\begin{align*}
    \log p(\rvu) = \log p(\rvv) + \log \Abs{\det\frac{\D \rvv}{\D \rvu}}
\end{align*}

Specifically, the change of variables from an 8-bit image to an image with pixel values in range [0, 1] is:
\begin{align*}
&\rvb_S^{(p)} = \frac{\rva_S^{(p)}}{256}\\
&\implies \log p(\rva_S) = \log p(\rvb_S) + \log \Abs{\det\frac{\D \rvb}{\D \rva}} \\
&\implies \log p(\rva_S) = \log p(\rvb_S) + \log \left(\frac{1}{256}\right)^{D_S} \\
&\implies \log p(\rva_S) = \log p(\rvb_S) - D_S \log 256\\
\implies
&\text{bpd}(\rva_S) 
= \frac{-\log p(\rva_S)}{D_S \log 2} \\
&= \frac{-(\log p(\rvb_S) - D_S \log 256)}{D_S \log 2} \\
&= \frac{-\log p(\rvb_S)}{D_S \log 2} + \frac{\log 256}{\log 2}\\
&= \text{bpd}(\rvx) + 8
\end{align*}
where $\text{bpd}(\rvx)$ is given from~\autoref{eq:bpd}.

\section{FID v/s Temperature}
Table \ref{tab:fid_vs_temp} lists the FID values of generated images from MRCNF models trained on CIFAR10, with different temperature settings on the Gaussian.
\begin{table}[!ht]
    \small
    \centering
    \begin{tabular}{|c@{\hskip 4pt}|@{\hskip 4pt}c@{\hskip 8pt}c@{\hskip 8pt}c@{\hskip 4pt}@{\hskip 4pt}c@{\hskip 8pt}c@{\hskip 8pt}c@{\hskip 4pt}|}
        \hline
        & \multicolumn{6}{c|}{\textbf{Temperature}} \\[2pt]
        & \textbf{1.0} & \textbf{0.9} & \textbf{0.8} & \textbf{0.7} & \textbf{0.6} & \textbf{0.5}
        \\[2pt]\hline
        \textbf{1-resolution CNF} & 138.82 & 147.62 & 175.93 & 284.75 & 405.34 & 466.16
        \\[2pt]
        \textbf{2-resolution MRCNF} & 89.55 & 106.21 & 171.53 & 261.64 & 370.38 & 435.17
        \\[2pt]
        \textbf{3-resolution MRCNF} & 88.51 & 104.39 & 152.82 & 232.53 & 301.89 & 329.12
        \\[2pt]
        \textbf{4-resolution MRCNF} & 92.19 & 104.35 & 135.58 & 186.71 & 250.39 & 313.39
        \\\hline
    \end{tabular}
    \caption{FID v/s temperature for \gls{msflow-image} models trained on CIFAR10.}
    \label{tab:fid_vs_temp}
\end{table}

\end{document}